\documentclass[12pt]{article}
\usepackage{geometry}
\usepackage{amssymb,amsmath,amsfonts,eurosym,geometry, graphicx,caption,color,setspace,sectsty,comment,footmisc,natbib,pdflscape,array,hyperref}
\usepackage{booktabs} 
\usepackage{subcaption} 
\usepackage{url}
\usepackage{pdflscape}
\usepackage{ragged2e}
\usepackage{multirow}
\usepackage{threeparttable}
\usepackage{float} 
\usepackage{indentfirst}
\usepackage{tabularx}
\usepackage{array}
\newcolumntype{Y}{>{\centering\arraybackslash}X}
\usepackage{setspace}
\usepackage{makecell}
\usepackage{afterpage}
\usepackage{xurl}
\usepackage{adjustbox} 
\usepackage{booktabs}  

\usepackage{siunitx}
\usepackage{caption}
\usepackage{adjustbox}

\newcolumntype{L}[1]{>{\raggedright\let\newline\\arraybackslash\hspace{0pt}}m{#1}}
\newcolumntype{C}[1]{>{\centering\let\newline\\arraybackslash\hspace{0pt}}m{#1}}
\newcolumntype{R}[1]{>{\raggedleft\let\newline\\arraybackslash\hspace{0pt}}m{#1}}

\begin{document}

\begin{titlepage}
\title{Shopping with a Platform AI Assistant: Who Adopts, When in the Journey, and What For}

\author{
  \makebox[\textwidth][c]{%
    \normalsize     
      Se Yan\thanks{Guanghua School of Management, Peking University, Beijing 10086, China. Email: \url{seyan@gsm.pku.edu.cn}.} \quad
      Han Zhong\thanks{Rotman School of Management, University of Toronto, Toronto, Ontario M5S 3E6, Canada. Email: \url{han.zhong@rotman.utoronto.ca}.} \quad
      Zemin (Zachary) Zhong\thanks{Rotman School of Management, University of Toronto, Toronto, Ontario M5S 3E6, Canada. Email: \url{zachary.zhong@rotman.utoronto.ca}. } \quad
      Wenyu Zhou\thanks{International Business School, Zhejiang University, Haining, Zhejiang 314400, China. Email: \url{wenyuzhou@zju.edu.cn}.} 
  }
  }

\date{\today}
\maketitle

\begin{abstract}

This paper provides some of the first large-scale descriptive evidence on how consumers adopt and use platform-embedded shopping AI in e-commerce. Using data on 31 million users of Ctrip, China’s largest online travel platform, we study “Wendao,” an LLM-based AI assistant integrated into the platform. We document three empirical regularities. First, adoption is highest among older consumers, female users, and highly engaged existing users, reversing the younger, male-dominated profile commonly documented for general-purpose AI tools. Second, AI chat appears in the same broad phase of the purchase journey as traditional search and well before order placement; among journeys containing both chat and search, the most common pattern is interleaving, with users moving back and forth between the two modalities. Third, consumers disproportionately use the assistant for exploratory, hard-to-keyword tasks: attraction queries account for 42\% of observed chat requests, and chat intent varies systematically with both the timing of chat relative to search and the category of products later purchased within the same journey. These findings suggest that embedded shopping AI functions less as a substitute for conventional search than as a complementary interface for exploratory product discovery in e-commerce.\\

\vspace{0in}\\
\end{abstract}
\setcounter{page}{0}
\thispagestyle{empty}
\end{titlepage}

\doublespacing


\section{Introduction}

The landscape of e-commerce is changing rapidly as major platforms embed LLM-powered assistants into the consumer shopping experience. Companies including Amazon, Alibaba, and Shopify have integrated generative AI into their core interfaces, and a recent Adobe survey reports that 38\% of U.S.\ consumers have used generative AI for online shopping.\footnote{\url{https://business.adobe.com/blog/generative-ai-powered-shopping-rises-with-traffic-to-retail-sites}} Yet we still know remarkably little about how shopping-specific AI is adopted and used at scale. Existing evidence on generative AI adoption comes primarily from general-purpose tools and suggests that early adopters are disproportionately younger, male, and more technically sophisticated \citep{draxler2023gender, otis2024global, chatterji2025people, yang2025adoption, bick2026rapid}. Whether the same pattern holds for platform-embedded shopping AI, where conversational assistance may reduce search and discovery frictions rather than reward technical sophistication, remains unclear. This leaves three basic questions unanswered: who adopts shopping-specific AI, where it fits in the purchase journey relative to traditional search, and what kinds of requests consumers delegate to it.

We study these questions using ``Wendao,'' an LLM-based shopping assistant embedded in Ctrip, China’s largest online travel platform. Our base sample consists of 31 million unique users who logged into the platform during a two-week period from July 10 to July 24, 2025. We define AI adopters as users who had initiated at least one chat with the assistant by July 10, 2025; under this definition, approximately 1.9 million users (6\%) are adopters. This setting provides two key empirical advantages. First, we observe AI conversations alongside rich purchase-journey data, including login activity, search queries, clicks, and transactions, allowing us to locate AI interactions relative to traditional search and booking events. Second, we observe rich user characteristics, including demographics, tenure, and loyalty status, which allow us to examine how the adoption of embedded shopping AI varies across consumer segments and prior platform relationships.

Our analysis establishes three empirical regularities. First, adoption is significantly higher among consumer segments that are often viewed as later adopters of new technologies, including older consumers (10.7\% for ages 50+ vs. 4.6\% for under 24) and female users. Adoption is also substantially higher among highly engaged existing users, indicating that take-up is concentrated among consumers with stronger prior ties to the platform. These patterns reverse the younger, male-dominated profile commonly documented for general-purpose AI tools and show that adoption of embedded shopping AI is strongest among users with established platform relationships. Second, AI chat appears in the same broad phase of the purchase journey as traditional product search and well before order placement. Among journeys containing both chat and search, the most common pattern is interleaving (53\%), with users moving back and forth between the two modalities. Third, attraction queries account for the largest share of observed user intent (42\%), and chat intent is closely aligned with downstream transaction categories: hotel chats are disproportionately followed by hotel orders, attraction chats by vacation orders, and so on. These patterns suggest that users rely on the assistant disproportionately for exploratory, harder-to-keyword tasks, and that observed chat activity is meaningfully connected to subsequent purchase needs.

These findings contribute to the emerging literature on generative AI in digital marketing. For theory, the reversal of standard adoption gradients suggests that platform-embedded AI may create value not  primarily as a productivity tool for technically sophisticated users, but by reducing interface frictions for consumers less comfortable with keyword-based search—helping them articulate needs, refine preferences, and navigate the platform. For empirical research, the strong association between prior platform engagement and adoption highlights the need for causal designs that distinguish selection into use from the effects of AI access itself, while the interleaving of chat and search implies that experiments randomizing AI access may also shift traditional search behavior. For practitioners, our results suggest that adoption is likely to be strongest among users with established platform relationships and that AI assistants should be optimized for exploratory discovery rather than transactional automation, while being designed to complement rather than simply replace conventional search. We discuss these implications further in Section~\ref{sec:conclusion}.

Our paper relates to three literatures. The first studies technology adoption and its demographic correlates \citep{rogers1962diffusion,  goldfarb2008internet}, including recent work on generative AI adoption \citep{draxler2023gender, otis2024global, handa2025economic, li2025from, tomlinson2025working, bick2026rapid, schubert2026household}. The second examines the role of AI in marketing, spanning both firm-side productivity applications \citep[e.g.][]{brynjolfsson2025generative} and consumer-facing assistants \citep[e.g.][]{lin2025towards,sun2025effect}. A closely related study in the same travel-platform setting examines, using a field experiment, how access to a reasoning versus non-reasoning AI assistant affects browsing and purchase outcomes \citep{yan2025impact}. Recent work has also begun to study how the adoption of general-purpose LLM reshapes online behavior more broadly \citep{padilla2025impact}. The third concerns consumer search and purchase decisions \citep[e.g.][]{ honka2019empirical, ursu2025sequential}. We contribute to these literatures by providing large-scale descriptive evidence on how platform-embedded shopping AI is adopted and used within a specific e-commerce setting, documenting patterns that differ in important ways from those reported for general-purpose AI tools.

\section{Context and Data}\label{sec:data}

We collaborate with Ctrip, China’s largest online travel platform. The platform allows consumers to search for and book hotels, flights, and ground transportation, as well as purchase attraction tickets and other trip-related products. Consumers typically interact with the platform through a standard e-commerce funnel involving keyword-based search, algorithmic ranking, filtering and sorting, product-detail browsing, and checkout. Ctrip has also deployed an LLM-powered conversational assistant, \textit{Wendao}, which is accessible through a persistent floating icon in the mobile app and through entry points on the web interface, allowing users to invoke the assistant while browsing search results. Users can open the assistant, submit travel-related questions, receive AI-generated responses, and navigate from the conversation to products on the platform. Relative to general-purpose AI tools, \textit{Wendao} has access to rich platform-specific information on travel products, including structured attributes, platform labels, user reviews, and other user-generated content. In addition, because it is built for this setting, the assistant is designed to better support travel-related discovery and comparison on the platform.

Our base sample consists of 31,142,353 users who logged into the app during a two-week window from July 10 to July 24, 2025. To study behavior around this focal sample, we observe user activity over a broader window from June 10 to August 15, 2025. We use this wider window because consumer activity in online travel is relatively sparse and episodic, so many users do not generate observable chat, search, or booking activity within any given two-week period. The broader window therefore provides a richer set of platform activities for tracing purchase journeys and relating chat behavior to downstream outcomes. Within this setting, we define an AI chat adopter as a user who had initiated at least one chat with \textit{Wendao} by July 10, 2025. Under this definition, 1,904,368 users (6.1\%) are adopters. Because \textit{Wendao} was launched before our observation window, many adopters first used the assistant prior to the period we observe. Moreover, because Ctrip is not a high-frequency-use platform for many consumers, many adopters do not generate observable chat activity during that window. Consequently, only a subset of adopters can be linked to observed chat transcripts: 188,647 adopters sent at least one chat request with an observable transcript.

We use four linked datasets, joined by an anonymized user identifier, to trace user activity on the platform. At the user level, we observe demographic and account attributes, including gender, age group, city, device type, platform tenure, and VIP status. At the hotel-search level, we observe timestamped search and browsing actions, including hotel search queries and clicks on hotel detail pages. At the chat level, we observe interactions with the AI assistant, including the full text of user requests and AI replies. The platform's internal algorithm also classifies each request into an intent category (e.g., hotel booking, travel planning, attractions, or customer service). Finally, we observe all bookings made on the platform, including order time and product category. Together, these linked datasets allow us to relate user characteristics to AI adoption and to situate chat interactions within broader search and purchase journeys on the platform.

\section{Who Adopts AI}\label{sec:who}
\begin{figure}[htbp!]
    \caption{AI chat adoption rates by demographic and account characteristics}
\centering\includegraphics[width=0.8\linewidth]{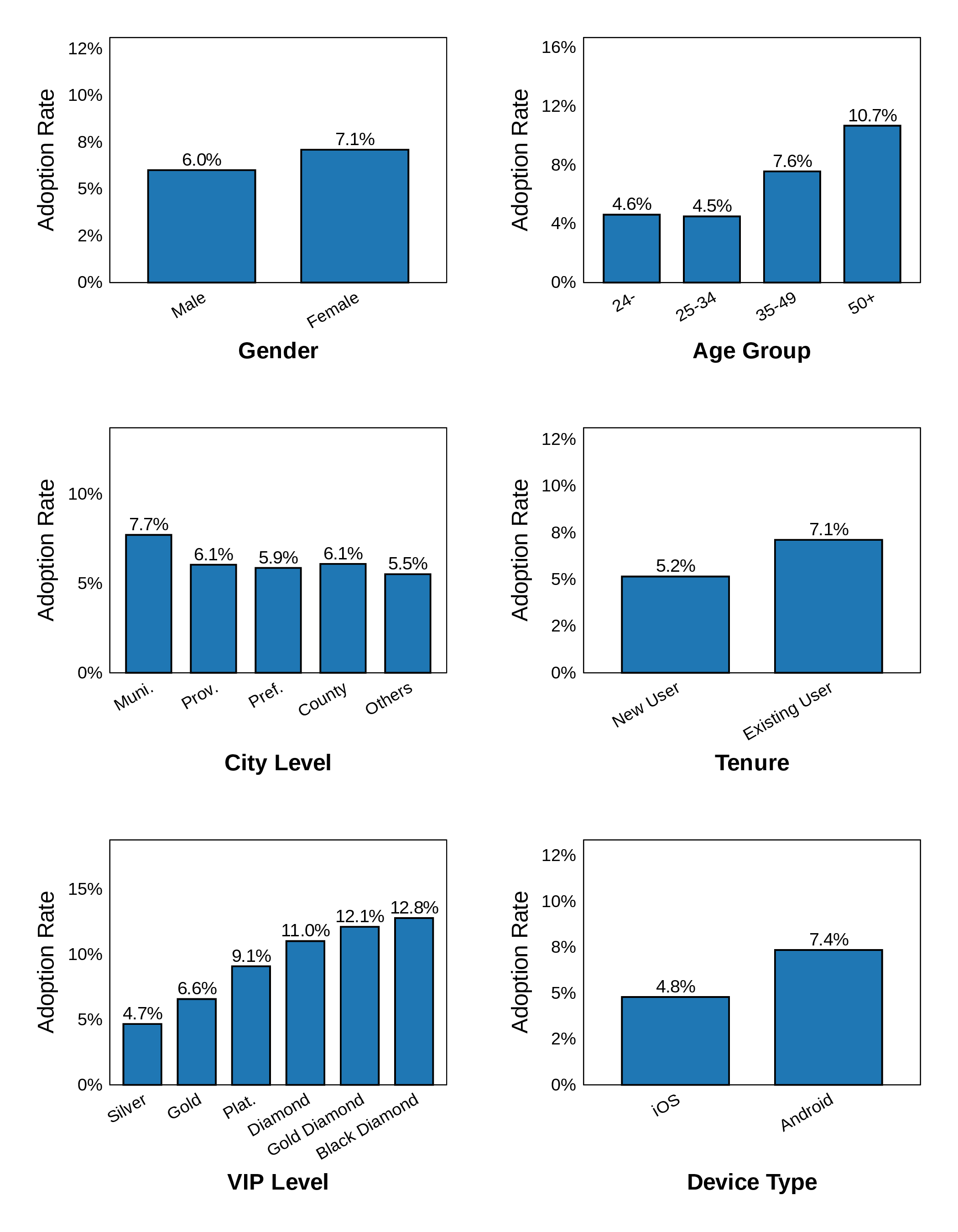}
    \label{fig:adoption_demographics}
    \begin{minipage}{0.8\linewidth}\setstretch{1}
    {\footnotesize \textit{Notes}: AI chat adoption is defined as initiating at least one chat session by July~10,~2025. Each panel reports adoption rates by user subgroup.}
    \end{minipage}
\end{figure}

In our data, 6.1\% of users had adopted the AI assistant by July 10, 2025. Although this rate is not directly comparable to self-reported survey estimates of generative AI use for online shopping (e.g., 38\% in a recent Adobe survey), it suggests that adoption of a specific platform-embedded shopping assistant remains far from universal, even as broader familiarity with AI shopping tools appears much higher. We examine these adoption patterns first through unconditional comparisons (Figure~\ref{fig:adoption_demographics}) and then through regression models estimating conditional associations (Figure~\ref{fig:adoption_regression}).

Figure~\ref{fig:adoption_demographics} presents unconditional adoption rates. Three patterns stand out. First, adoption increases monotonically with age (10.7\% for ages 50+ vs.\ 4.6\% for users under 24), reversing the younger, male-dominated profile commonly documented for general-purpose AI tools \citep{draxler2023gender, otis2024global, bick2026rapid}. Second, female users adopt at a modestly higher rate than male users (7.1\% vs.\ 6.0\%)\footnote{The female and male adoption rates are computed among users with observed gender, so they need not average to the full-sample adoption rate. In the gender-known subsample, the pooled adoption rate is 6.58\%; the full-sample rate is lower because users with missing gender have a lower adoption rate (3.43\%).}, again contrasting with the male-skewed pattern often found in broader AI adoption \citep{draxler2023gender}. Third, adoption rises sharply with loyalty tier (12.8\% for Black Diamond VIP users vs.\ 4.7\% for Silver users). Longer-tenure users also exhibit higher adoption rates. Interestingly, adoption is also higher among Android users than iOS users. To the extent that iOS is often associated with more expensive devices, this pattern runs against a simple premium-user or technical-sophistication account of early AI adoption, though operating system is an imperfect proxy for income and may also capture broader differences in user composition. Comparing adopters and non-adopters directly (Appendix Table~\ref{tab:sumstat_adopter_nonadopter}) confirms these demographic gradients and shows that adopters are substantially more engaged on the platform: they record more than double the login minutes (797 vs.\ 355) and nearly twice the number of search queries (150 vs.\ 67). Together, these patterns indicate that AI adoption on the platform is concentrated among older, female, and more strongly attached users, and is closely associated with overall platform engagement.

To assess which factors are associated with adoption, we regress an indicator for AI chat adoption on demographic, account, and engagement variables using a linear probability model, a logit model, and a probit model. Appendix Figure~\ref{fig:adoption_regression} plots the estimated marginal effects, which are consistent across all three specifications. Several patterns from the unconditional comparisons survive after controlling for behavioral variables. Age remains the strongest demographic predictor: relative to the under-24 baseline, users aged 50+ are roughly 4--5 percentage points more likely to adopt, a large effect against the 6.1\% base rate. Being female is associated with an increase of approximately 1.5 percentage points. The iOS device indicator enters negatively (about $-$1.5 pp). VIP status and star rating retain modest positive effects after controlling for engagement, suggesting that loyalty captures something beyond mere usage intensity.

Among behavioral variables, login sessions and average session duration show the largest positive effects, while total orders contributes little incremental predictive power once engagement intensity is controlled, suggesting that the depth of \emph{search} behavior, rather than purchasing per se, distinguishes adopters.\footnote{The negative coefficient on total login minutes in Figure~\ref{fig:adoption_regression} is an artifact of collinearity: because total login minutes is approximately the product of login sessions and average minutes per session, including all three simultaneously produces unstable individual coefficients. The variable-importance ranking in Appendix Figure~\ref{fig:adoption_rf_importance}, which is immune to collinearity, confirms that all three engagement measures are individually informative.}

A random forest classifier reinforces these findings (Appendix Figure~\ref{fig:adoption_rf_importance}): engagement variables (login minutes, sessions, session duration) dominate predictive importance, followed by age. Adding engagement to demographics raises the random forest AUC (area under the ROC curve, where 0.5 is chance and 1.0 is perfect discrimination) from 0.635 to 0.962 (Appendix Table~\ref{tab:model_performance}), confirming that engagement contains rich nonlinear signals that demographics alone miss.

One important caveat: the strong predictive power of engagement should not be read causally. Heavy users mechanically encounter the AI entry point more often, so the association can reflect greater exposure rather than greater propensity. Disentangling the two would require exogenous variation in feature visibility, which is beyond this descriptive analysis.

The parametric and nonparametric results tell a consistent story. The adoption profile is strikingly different from that documented for general-purpose AI tools: adoption rises with age, is higher among female users, and is concentrated among users with stronger prior ties to the platform. At the same time, the single strongest predictor of adoption is the intensity of existing platform engagement, although, as noted above, this association may partly reflect greater exposure to the AI entry point rather than a higher underlying propensity to adopt.

%



\section{When Consumers Use AI}\label{sec:when}

Having established \textit{who} adopts the AI assistant, we now turn to \textit{when} it is used within the purchase journey. In particular, how does AI chat relate to traditional search in the path to purchase? Does it tend to appear early as a discovery tool, or does it arise later in journeys that are already underway? We address this in two steps: first, we establish where AI chat occurs relative to other journey events on average; then, we examine whether it occurs before, after, or alongside traditional search.

We define a customer journey from login sessions in two steps. First, we construct sessions from login and logout timestamps and stitch adjacent sessions into a provisional journey unless the gap between one session's logout and the next session's login is at least 48 hours. Second, within each provisional journey, if multiple sessions contain orders, we split the journey between two adjacent order-containing sessions whenever the inactivity gap between the earlier session's logout and the later session's login exceeds 4 hours.\footnote{This second step is intended to capture the idea that a purchase-oriented shopping episode often ends after an order followed by a meaningful pause. We therefore view the resulting journey as an operational approximation to a latent shopping episode.} In the online appendix, we show that the main qualitative patterns are similar under alternative definitions: an inactivity-only definition that omits the second step and more permissive order-gap thresholds. Under our baseline definition, we identify over 253 million journeys during the observation window (Appendix Table~\ref{tab:summary_stats}). The median journey lasts 7 hours, spans 2 login sessions, and involves 11 minutes of total login time. Most journeys are brief: the median contains zero chat sessions, zero clicks, and zero orders. At the same time, the distributions are strongly right-skewed, reflecting a long tail of more involved travel-planning episodes.\footnote{Appendix Figure~\ref{fig:journey} illustrates journey heterogeneity by plotting the timelines of ten randomly selected users over a 30-day window.}

\begin{figure}[htbp!]
    \caption{Density of event types over the course of the customer journey}
\centering\includegraphics[width=0.9\linewidth]{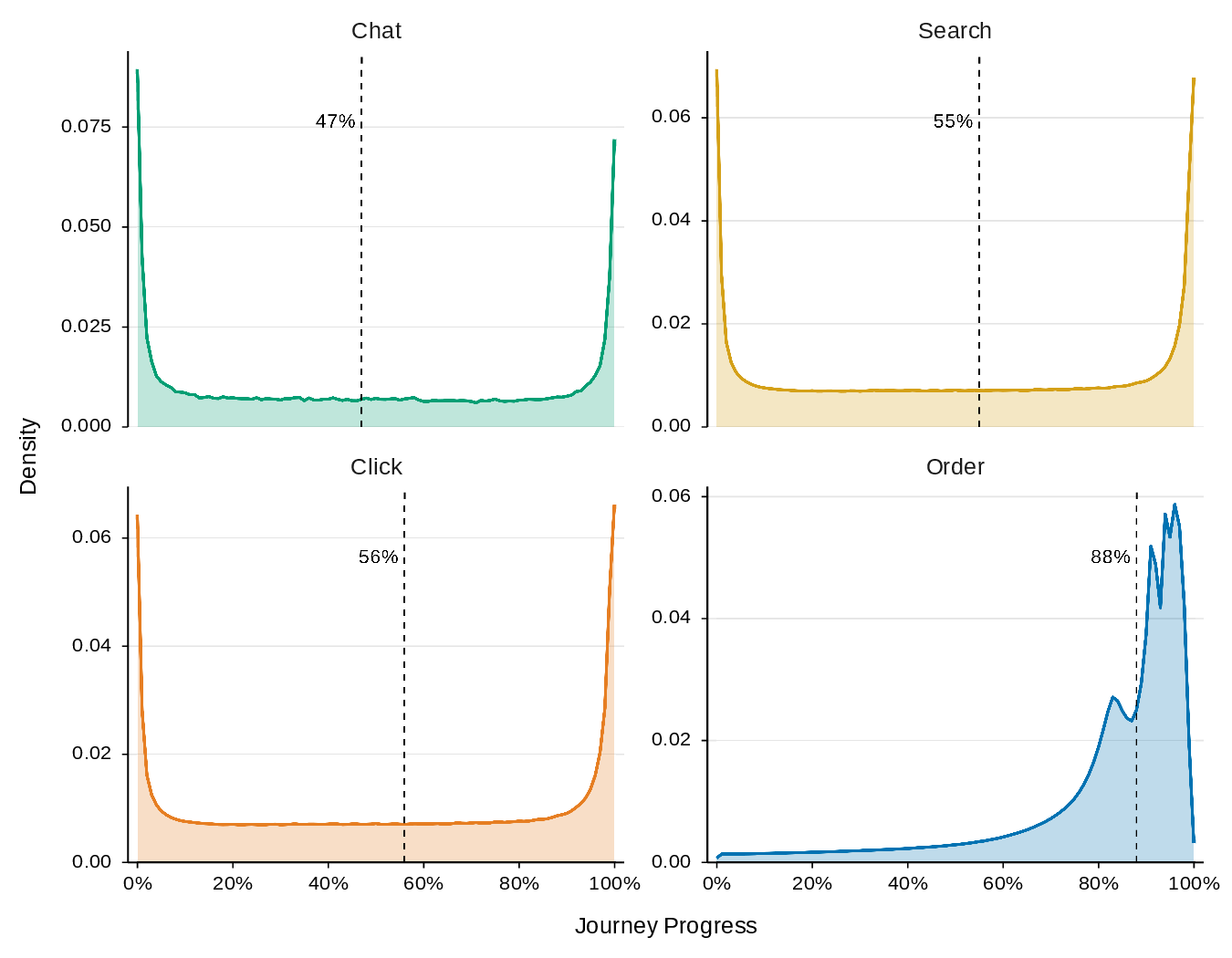}
    \label{fig:progression_density}
    \begin{minipage}{0.9\linewidth}\setstretch{1}
    {\footnotesize \textit{Notes:} Each panel shows the kernel density of an event type over normalized journey progress (0\%--100\% of total journey duration). The dashed line marks the median event position.
    }
    \end{minipage}
\end{figure}

To understand where chat fits in the purchase funnel, we normalize each event's timestamp to a $[0\%, 100\%]$ scale representing its relative position within the journey and plot kernel density estimates for each event type. Figure~\ref{fig:progression_density} displays the results. The median AI chat event occurs at 47\% of journey progress, closer to the median for hotel searches (55\%) and clicks (56\%) than to orders, whose median falls at 88\%. The chat density follows a broadly similar distributional shape to search and clicks, with spikes near both the beginning and end of the journey and a relatively smooth distribution in between, though it is shifted somewhat earlier overall. Order density, by contrast, is concentrated near the journey's end. Recentering the timeline on the chat event itself confirms that searches and clicks are tightly centered around chat, while orders are displaced well to the right (Appendix Figure~\ref{fig:norm_chat}). This pattern indicates that AI chat is primarily a pre-purchase activity, occurring in a similar phase of the journey as active search and browsing rather than as post-booking support. Recentering the timeline on the order event yields the same qualitative conclusion: chat, search, and clicks generally occur before order placement (Appendix Figures~\ref{fig:norm_order_wc} and~\ref{fig:norm_order_nc}). Appendix Figures~\ref{fig:progression_density_48h}, ~\ref{fig:progression_density_48h_8h} and~\ref{fig:progression_density_48h_12h} show very similar patterns under alternative journey definitions, including an inactivity-only definition that omits the order-based splitting step and definitions using more permissive order-gap thresholds of 8 and 12 hours.

The average timing result masks important heterogeneity in how chat is positioned relative to traditional search. We classify journeys containing at least one chat session into four mutually exclusive groups: \textit{chat-before-search} journeys (all chat events precede all search events), \textit{chat-after-search} journeys (all chat events follow all search events), \textit{chat-between-search} journeys (chat and search events are interspersed), and \textit{chat-only} journeys (no hotel search occurs). These patterns may reflect distinct functional roles for AI: in \textit{chat-before-search} journeys, the assistant may serve as an entry point for exploration; in \textit{chat-after-search} journeys, it may serve as a refinement tool invoked after keyword search; and in \textit{chat-only} journeys, it may substitute for hotel search within the journey.

Table~\ref{tab:chat_sequence} reports the results. Chat-only journeys account for the largest share (42\%), indicating that a substantial fraction of chat use occurs without any accompanying hotel search. Among all chat journeys, chat-between-search accounts for 31\%, followed by chat-before-search (15\%) and chat-after-search (12\%). Conditional on a journey containing both chat and search, however, the chat-between-search pattern dominates, accounting for 53\% of such journeys, compared with 26\% for chat-before-search and 21\% for chat-after-search. Interleaved journeys are substantially longer (median 95 hours vs.\ roughly 52 for the other two chat-and-search groups), involve far more hotel searches (150 vs.\ 21--24), and exhibit the highest conversion rate (43\% vs.\ 39\% for chat-before-search and 31\% for chat-after-search). Chat-only journeys, by contrast, are the shortest (median 19 hours) and have the lowest conversion rate (18\%), consistent with a more lightweight or exploratory usage pattern. Taken together, the data suggest that chat can serve as an entry point for search (chat-before-search), as a refinement tool following search (chat-after-search), or as a standalone discovery channel (chat-only), but when both modalities are present it is most often interleaved with search, consistent with users going back and forth between the two modalities.  

One caveat is that longer journeys with more actions mechanically create more opportunities for interleaving: a short journey with one chat and one search can only be classified as chat-before-search or chat-after-search, never chat-between-search. The higher interleaving share among journeys containing both modalities may therefore partly reflect the dominance of longer, more engaged journeys in that subsample. To assess whether this is the case, Appendix Table~\ref{tab:chat_sequence_session} repeats the analysis at the session level, where the scope for mechanical interleaving is more limited. Chat-between-search remains the most prevalent pattern among sessions containing both chat and search, though its share is notably smaller than at the journey level, consistent with back-and-forth use of chat and search often unfolding across multiple sessions rather than within a single login. The gap in conversion rates across groups is even more pronounced at the session level, suggesting that the association between interleaving and purchase is unlikely to be driven solely by journey length.

A particularly revealing dimension is Panel~C of Table~\ref{tab:chat_sequence}, which reports the distribution of chat topics across the four groups. (The platform's internal algorithm classifies each chat session by topic; we describe the full intent taxonomy in Section~\ref{sec:how}.) Relative to the other groups, chat-before-search journeys are somewhat more likely to involve travel planning queries (9.8\%), consistent with users turning to the assistant earlier in the journey for broader exploration. Chat-between-search journeys have the highest hotel intent share (30.4\%), consistent with users toggling between keyword search and conversational AI to refine hotel-specific decisions. Chat-after-search journeys are the most attraction-heavy (48.6\%), suggesting that users who have already searched for hotels turn to AI for complementary information about destinations and activities. Chat-only journeys, by contrast, are the least hotel-oriented and relatively more concentrated in other transportation, customer service, and residual topics, suggesting that standalone chat use is often less tightly linked to hotel search. These patterns are robust to alternative journey definitions, including a 48-hour inactivity-only definition and more permissive order-gap thresholds of 8 and 12 hours (Appendix Tables~\ref{tab:chat_sequence_48h}, ~\ref{tab:chat_sequence_48h_8h} and~\ref{tab:chat_sequence_48h_12h}), and the same topic ordering appears at the session level (Appendix Table~\ref{tab:chat_sequence_session}, Panel~C).

\begin{table}[htbp!]
\centering
\caption{Journey Characteristics by Relative Timing of Chat and Search}
\label{tab:chat_sequence}
\footnotesize
\begin{threeparttable}
\begin{tabular}{lcccc}
\toprule
 & Chat-Before-Search & Chat-Between-Search & Chat-After-Search & Chat-Only \\
\midrule
\multicolumn{5}{l}{\textit{\textbf{Panel A: Prevalence}}} \\
Share of journeys (\%) & 15.076 & 30.916 & 12.411 & 41.596 \\
\addlinespace[0.35em]
\multicolumn{5}{l}{\textit{\textbf{Panel B: Journey characteristics}}} \\
Median journey length (hours) & 51.809 & 94.837 & 51.970 & 18.405 \\
Mean login sessions & 11.313 & 28.625 & 10.538 & 5.582 \\
Mean chat sessions & 1.377 & 1.566 & 1.358 & 1.411 \\
Mean hotel searches & 24.365 & 150.029 & 21.433 & 0.000 \\
Mean clicks & 11.504 & 68.371 & 9.532 & 0.000 \\
Mean orders & 0.562 & 0.666 & 0.463 & 0.223 \\
Conversion rate (\%) & 39.382 & 43.166 & 31.130 & 18.040 \\
\addlinespace[0.35em]
\multicolumn{5}{l}{\textit{\textbf{Panel C: Chat intent distribution (\%)}}} \\
Attraction & 44.210 & 41.881 & 48.481 & 39.873 \\
Hotel & 17.927 & 30.388 & 17.681 & 7.062 \\
Travel Planning & 9.837 & 6.075 & 5.390 & 8.031 \\
Customer Service & 4.871 & 3.763 & 5.541 & 6.930 \\
Flight & 1.369 & 0.589 & 0.919 & 3.243 \\
Other Transportation & 6.233 & 4.870 & 5.615 & 12.692 \\
Other & 15.552 & 12.434 & 16.373 & 22.169 \\
\bottomrule
\end{tabular}
\begin{tablenotes}[flushleft]
\item \textit{Notes:} Journeys are classified into four groups containing at least one AI chat session: \textit{Chat-Before-Search} (all chat events precede all hotel search events), \textit{Chat-After-Search} (all chat events follow all hotel search events), \textit{Chat-Between-Search} (chat and search events are interleaved), and \textit{Chat-Only} (chat but no hotel search). Panel~A reports group shares, Panel~B journey-level characteristics, and Panel~C chat intent distributions. Intent shares sum to 100\% within each column, up to rounding.
\end{tablenotes}
\end{threeparttable}
\end{table}

\section{How Consumers Use AI}\label{sec:how}

We next examine how consumers use the AI assistant in practice. The platform assigns each chat session to one of seven intent categories using its internal classifier: Attraction, Hotel, Travel Planning, Consumer Support, Flight, Other Transportation, and Other. Appendix Table~\ref{tab:example_conversations} presents illustrative conversations for each category, translated from Chinese.

Before turning to intent composition, we briefly characterize the intensive margin of AI engagement. Among the 188,647 adopters with observable chat transcripts, the median user initiates a chat in less than 4\% of login sessions (Appendix Table~\ref{tab:sumstat_ai_assistant}). Sessions are brief and typically concentrated on a single query: the median session contains one request, the median user message is 8 characters long, and the median AI reply is 1,145 characters. Roughly 79\% of sessions use the assistant's reasoning mode, an extended-inference feature enabled by default. Conditional on adoption, the number of sessions per user and requests per session vary little across gender, age, city tier, VIP level, and device type, although session duration differs somewhat, with older users and Android users engaging in longer sessions (Appendix Figures~\ref{fig:avg_sessions_per_user},~\ref{fig:avg_session_duration}, and~\ref{fig:avg_requests_per_session}). This relative homogeneity on the intensive margin contrasts with the pronounced demographic differences in adoption documented in Section~\ref{sec:who}. The contrast suggests that the main margin of differentiation is whether consumers try the assistant at all, rather than how intensively they use it once they adopt.

Section~4 showed that chat intent varies with how users combine conversational AI and traditional search within a journey. We now step back and summarize the overall distribution of request types across all chat interactions. Attraction queries dominate overall, accounting for 42\% of all requests, followed by hotels (18\%), other (17\%), and travel planning (7\%). Customer support and transportation queries are relatively rare (5\% and 8\%, respectively). Intent varies meaningfully by age: younger users (under 24) are far more likely to seek customer support (11\% vs.\ 3\% for ages 50+), while older users disproportionately ask about hotels (21\% vs.\ 15\%). Across other dimensions, including gender, city tier, VIP level, device type, and tenure, the intent distribution is fairly stable (Appendix Table~\ref{tab:intent_share_by_group}). 

\begin{figure}[htbp!]
    \caption{Temporal distribution of orders relative to chat sessions by chat intent}
\centering\includegraphics[width=0.85\linewidth]{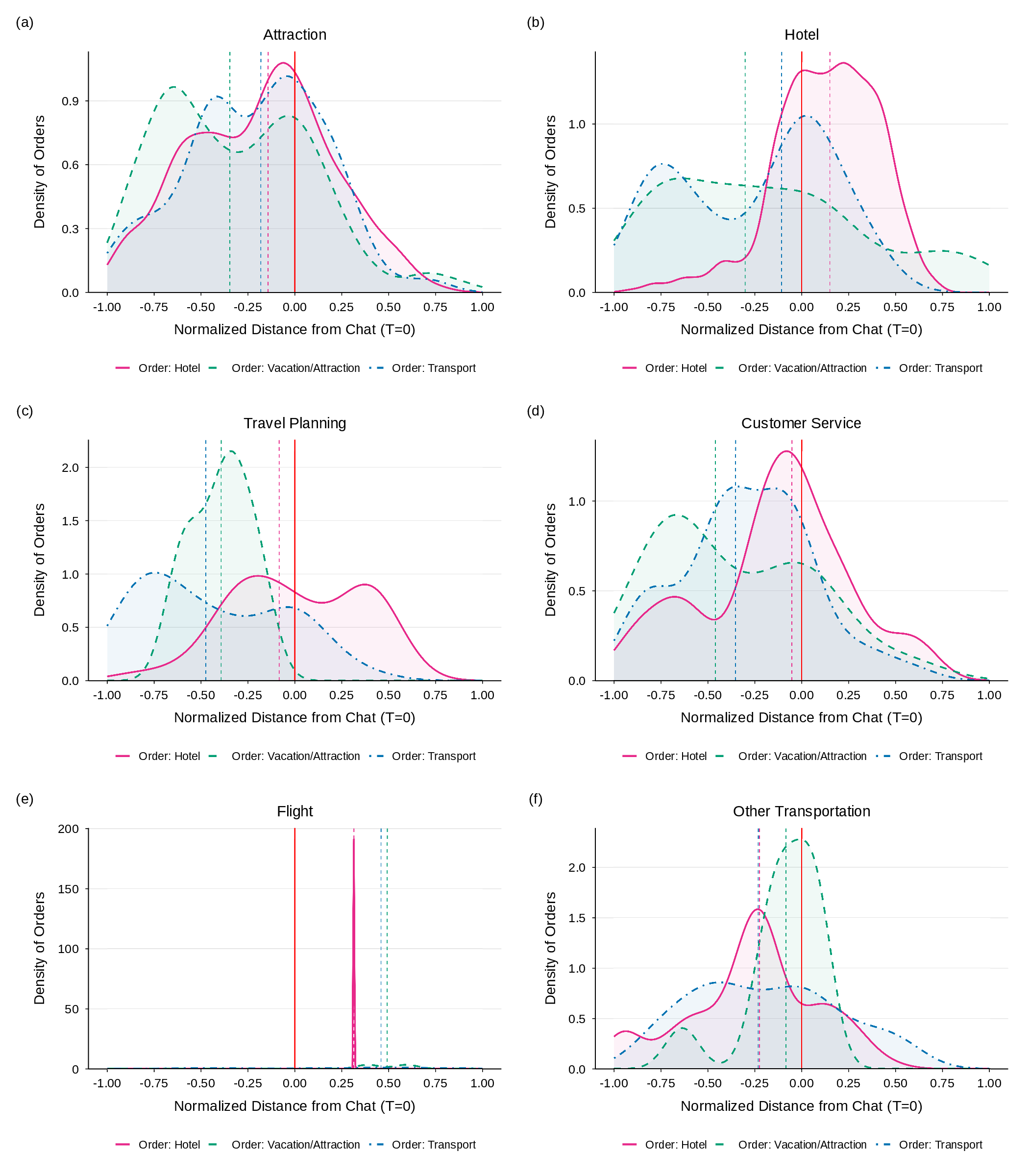}
    \label{fig:order_density_by_intent}
    \begin{minipage}{0.95\linewidth}\setstretch{1}
    {\footnotesize \textit{Notes:} Each panel shows the kernel density of orders relative to the chat event ($T=0$) for a given chat intent category. The horizontal axis reports normalized distance from the chat event; negative values indicate orders placed before the chat and positive values indicate orders placed after. Within each panel, densities are shown separately by order type. The dashed line marks the median relative order timing. }
    \end{minipage}
\end{figure}

As shown in Section~4, the intent mix differs systematically across chat--search sequence patterns (Table~\ref{tab:chat_sequence}, Panel~C). Here we emphasize a broader implication of those results: attraction-related chat is not confined to the earliest, purely exploratory stage of the journey. Its share remains broadly stable across journey progress quintiles (Appendix Figure~\ref{fig:intent_by_progress}), and Figure~\ref{fig:order_density_by_intent} (panel a) shows that orders occur both before and after attraction-related chats. The prominence of attraction queries is notable given that Ctrip is primarily known as a hotel and flight booking platform. These patterns suggest that users rely on conversational AI for open-ended destination and activity questions that are difficult to express as keywords, not only for initial inspiration but also for ongoing destination discovery alongside search and booking.

Figure~\ref{fig:order_density_by_intent} examines whether chat intent aligns with subsequent purchasing behavior by plotting the kernel density of orders relative to the chat event ($T=0$), separately by intent category and order type. A clear alignment emerges. When users ask about hotels (panel b), hotel orders cluster sharply just after the chat; when users ask about attractions (panel a), vacation and attraction orders become relatively more prominent. Travel planning chats (panel c) are followed by a mix of order types, consistent with their broad, exploratory nature. Consumer support chats (panel d) show orders concentrated before the chat, suggesting that these interactions often follow rather than precede a transaction. Flight and transportation panels (e, f) show corresponding spikes in transport orders. These patterns indicate that AI chat intent is not idle browsing but is correlated with the type of products users ultimately purchase.





%


\section{Conclusion}\label{sec:conclusion}

Using data on 31 million users of Ctrip, China’s largest online travel platform, we document three sets of stylized facts about the adoption and usage of an LLM-based shopping assistant. First, adoption is highest among older consumers, female users, and existing rather than new platform users, reversing the younger, male-skewed profile often documented for general-purpose AI tools. At the same time, the single strongest predictor of adoption is the intensity of prior platform engagement. Second, AI chat occurs in the same broad phase of the purchase journey as traditional search and generally before order placement. A substantial share of chat use occurs without any accompanying hotel search, but when both modalities are present, the most common pattern is interleaving, consistent with users moving back and forth between chat and search within the same journey. Third, users disproportionately turn to the assistant for exploratory, hard-to-keyword tasks: attraction queries account for 42\% of all chat requests, and chat intent varies systematically with both chat--search sequence patterns and the category of products subsequently purchased.

These findings offer several implications for academic researchers. For theory, our results suggest that existing frameworks of technology adoption, which emphasize technical literacy and innovativeness as drivers of early adoption, may not apply when AI is embedded within a domain-specific commercial interface. The reversal of the standard age and gender gradients points toward a search-cost-reduction mechanism: users who face higher costs navigating traditional keyword interfaces may benefit most from conversational AI, making them more likely to adopt rather than less. Formalizing this mechanism is a promising direction for theoretical work. For empirical research, our descriptive findings can inform the design and interpretation of causal studies. The strong association between engagement and adoption highlights the need for identification strategies that separate exposure from preference, such as randomized variation in the visibility or placement of AI features. The interleaving of chat and search suggests that experiments randomizing AI access may affect not only direct outcomes but also traditional search behavior, an important consideration for experimental design. The intent-to-purchase correlation documented in Figure~\ref{fig:order_density_by_intent} provides a natural starting point for studying whether AI chat causally shifts purchase composition or simply reflects pre-existing intent.

For platform managers, our results highlight two design priorities. First, since an important margin appears to be initial trial rather than continued engagement, strategies that encourage first-time interactions, such as contextual prompts during active search, may be more effective than features aimed at deepening usage among existing adopters. Second, the dominance of exploratory, open-ended queries (attractions, travel planning) suggests that AI assistants should be optimized for discovery and inspiration rather than narrowly for transactional automation. For policymakers concerned with equitable access to AI-enhanced services, the finding that older consumers and users not typically viewed as early adopters of general-purpose AI are the most active adopters in this setting provides a counterpoint to concerns that generative AI will primarily benefit younger, digitally fluent populations, though this pattern may be specific to platform-embedded contexts where AI reduces rather than increases complexity.
\section*{Funding and Competing Interests}

All authors certify that they have no affiliations with or involvement in any organization or entity with any financial interest or non-financial interest in the subject matter or materials discussed in this manuscript. The authors have no funding to report.

\bibliographystyle{econ-a}
\bibliography{ref}

\clearpage
\appendix

\setcounter{page}{1}
\setcounter{table}{0}
\setcounter{figure}{0}
\renewcommand{\thetable}{A\arabic{table}}
\renewcommand{\thefigure}{A\arabic{figure}}
\renewcommand{\thesection}{A\arabic{section}}

\begin{center}
\vspace*{2cm}
{\LARGE \textbf{Online Appendix to}}\\[12pt]
{\Large \textbf{Shopping with a Platform AI Assistant: Who Adopts, When in the Journey, and What For}}\\[24pt]
{\large March 2026}\\[6pt]
\vspace*{2cm}
\end{center}

\clearpage
\section{Additional Tables and Figures}


\begin{table}[htbp]
\centering
\caption{Summary statistics by AI chat adoption status}
\label{tab:sumstat_adopter_nonadopter}
\small
\setlength{\tabcolsep}{4pt}
\begin{threeparttable}
\resizebox{\textwidth}{!}{%
\begin{minipage}{\textwidth}
\centering
\renewcommand{\arraystretch}{1.5}
\begin{tabular}{lcccccccc}
\toprule
& \multicolumn{3}{c}{AI chat adopters} & \multicolumn{3}{c}{AI chat non-adopters} & \multicolumn{2}{c}{Difference} \\
\cmidrule(lr){2-4}\cmidrule(lr){5-7}\cmidrule(lr){8-9}
Variable & N & Mean & S.D. & N & Mean & S.D. & Diff. & $t$-stat \\
\midrule
\multicolumn{9}{l}{\textit{\textbf{Panel A: Demographics and account status}}}\\
Female                               & 1,904,368  & 0.516 & 0.500 & 29,237,985 & 0.440 & 0.496 & 0.076  & 204.492 \\
Age group                            & 1,904,368  & 2.616 & 1.096 & 29,237,985 & 2.186 & 1.180 & 0.430  & 521.849 \\
New user indicator                   & 1,904,368  & 0.430 & 0.495 & 29,237,985 & 0.516 & 0.500 & $-$0.085 & $-$230.466 \\
iOS device indicator                 & 1,904,368  & 0.379 & 0.485 & 29,237,985 & 0.490 & 0.500 & $-$0.112 & $-$306.670 \\
VIP status (levels 4--6)             & 1,904,368  & 0.123 & 0.328 & 29,237,985 & 0.062 & 0.242 & 0.060  & 249.793 \\
High star rating (4--5)              & 1,904,368  & 0.305 & 0.460 & 29,237,985 & 0.249 & 0.433 & 0.056  & 163.219 \\
City level                           & 1,904,368  & 2.167 & 0.892 & 29,237,985 & 2.172 & 0.939 & $-$0.005 & $-$7.814 \\
\addlinespace[0.35em]
\multicolumn{9}{l}{\textit{\textbf{Panel B: Platform engagement}}}\\
Total clicks                         & 1,904,368  & 66.507 & 184.512 & 29,237,985 & 28.174 & 89.014 & 38.333 & 284.546 \\
Total search queries                 & 1,904,368  & 150.486 & 619.324 & 29,237,985 & 66.772 & 347.792 & 83.714 & 184.646 \\
Total login minutes                  & 1,904,368  & 796.904 & 1,335.474 & 29,237,985 & 354.714 & 660.673 & 442.191 & 453.331 \\
Number of login sessions             & 1,904,368  & 59.653 & 68.515 & 29,237,985 & 32.788 & 44.218 & 26.864 & 533.890 \\
Average minutes per session          & 1,904,368  & 11.798 & 8.724 & 29,237,985 & 9.290 & 11.614 & 2.508  & 375.627 \\
\addlinespace[0.35em]
\multicolumn{9}{l}{\textit{\textbf{Panel C: Purchasing activity}}}\\
Total orders (any type)              & 1,904,368  & 4.015 & 10.159 & 29,237,985 & 2.484 & 5.588 & 1.530  & 205.858 \\
Hotel orders                         & 1,904,368  & 1.588 & 6.615 & 29,237,985 & 0.900 & 2.948 & 0.689  & 142.716 \\
Flight orders                        & 1,904,368  & 0.689 & 2.240 & 29,237,985 & 0.484 & 2.150 & 0.205  & 122.708 \\
Train orders                         & 1,904,368  & 1.237 & 3.429 & 29,237,985 & 0.956 & 2.938 & 0.281  & 110.498 \\
Vacation package orders              & 1,904,368  & 0.048 & 0.389 & 29,237,985 & 0.017 & 0.153 & 0.031  & 110.669 \\
Attraction/activity orders           & 1,904,368  & 0.043 & 0.923 & 29,237,985 & 0.013 & 0.372 & 0.030  & 44.694  \\
Ticket orders                        & 1,904,368  & 0.410 & 5.288 & 29,237,985 & 0.115 & 1.597 & 0.294  & 76.577  \\
\bottomrule
\end{tabular}
\small
\begin{tablenotes}[flushleft]
\item \textit{Notes}: This table reports means and standard deviations for AI chat adopters and non-adopters. Diff.\ is the difference in means between adopters and non-adopters. The reported $t$-statistic tests equality of group means. 
\end{tablenotes}
\end{minipage}%
}
\end{threeparttable}
\end{table}


\begin{figure}[H]
    \caption{Estimated marginal effects on AI chat adoption from linear probability, logit, and probit models}
\centering\includegraphics[width=0.7\linewidth]{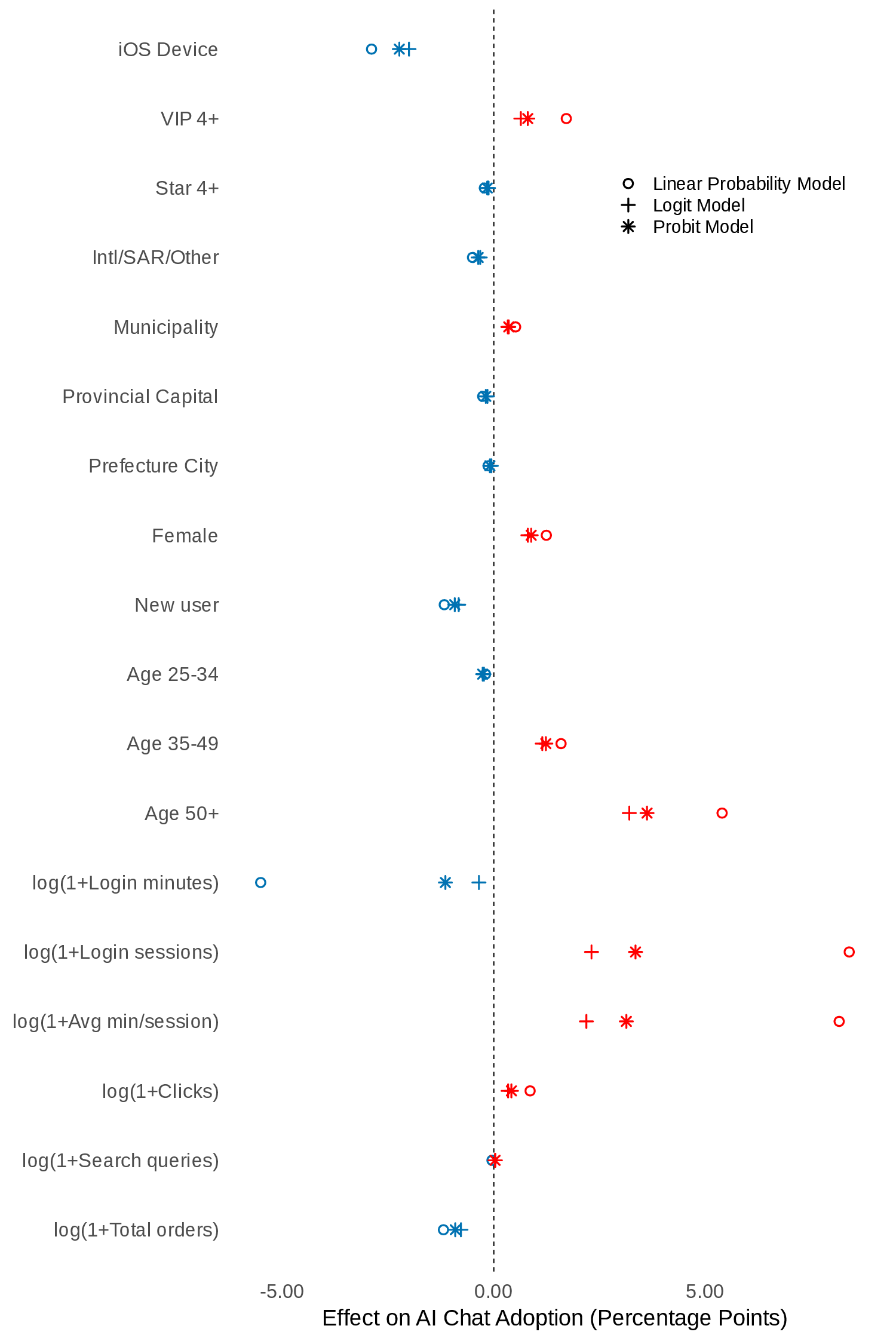}
    \label{fig:adoption_regression}
    \begin{minipage}{0.95\linewidth}\setstretch{1}
    {\footnotesize \textit{Notes}: This figure reports estimated marginal effects on AI chat adoption from the linear probability, logit, and probit models, expressed in percentage points. The covariates are listed on the vertical axis. The vertical dashed line marks zero, and the horizontal axis shows the estimated effect in percentage points.}
    \end{minipage}
\end{figure}

\begin{figure}[htbp]
    \caption{Permutation-based variable importance from a random forest model of AI chat adoption}
\centering\includegraphics[width=0.7\linewidth]{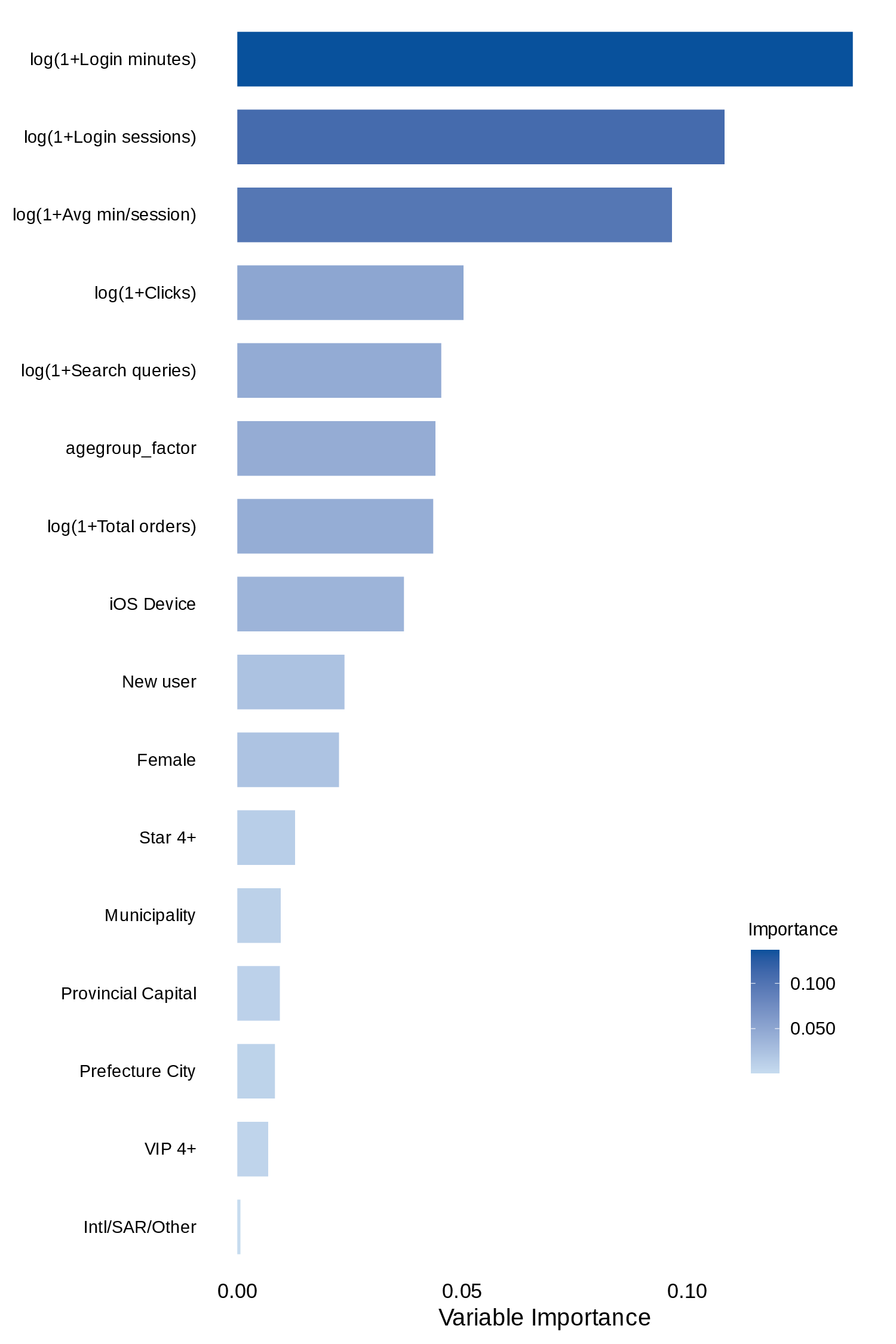}
    \label{fig:adoption_rf_importance}
    \begin{minipage}{1\linewidth}\setstretch{1}
    {\footnotesize \textit{Notes}: This figure reports permutation-based variable-importance scores from a random forest model of AI chat adoption. Each bar shows the decrease in model accuracy when the corresponding feature is randomly permuted, with larger values indicating greater predictive importance. Features are listed on the vertical axis.}
    \end{minipage}
\end{figure}

\newpage

\begin{table}[htbp]
\centering
\caption{Predictive Performance of Adoption Models}
\label{tab:model_performance}
\begin{threeparttable}
\begin{tabular*}{0.90\textwidth}{@{\extracolsep{\fill}}lcc@{}}
\toprule\small
Specification & LPM $R^2$ & RF AUC \\
\midrule
Demographics only          & 0.014 & 0.635 \\[3pt]
Demographics + Engagement  & 0.036 & 0.962 \\
\bottomrule
\end{tabular*}
\begin{tablenotes}[flushleft]
\small
\item \textit{Notes:} This table reports the predictive performance of AI chat adoption models under two sets of covariates. LPM $R^2$ is the coefficient of determination from a linear probability model, and RF AUC is the area under the ROC curve from a random forest classifier. The demographics-only specification includes female, age group, new user, iOS device, VIP status, high star rating, and city level. The demographics-plus-engagement specification additionally includes total login minutes, number of login sessions, average minutes per session, total clicks, total search queries, and total orders.
\end{tablenotes}
\end{threeparttable}
\end{table}

\newpage

\begin{figure}[H]
    \caption{Illustrative customer journey timelines for ten randomly selected users}
\centering\includegraphics[width=1\linewidth]{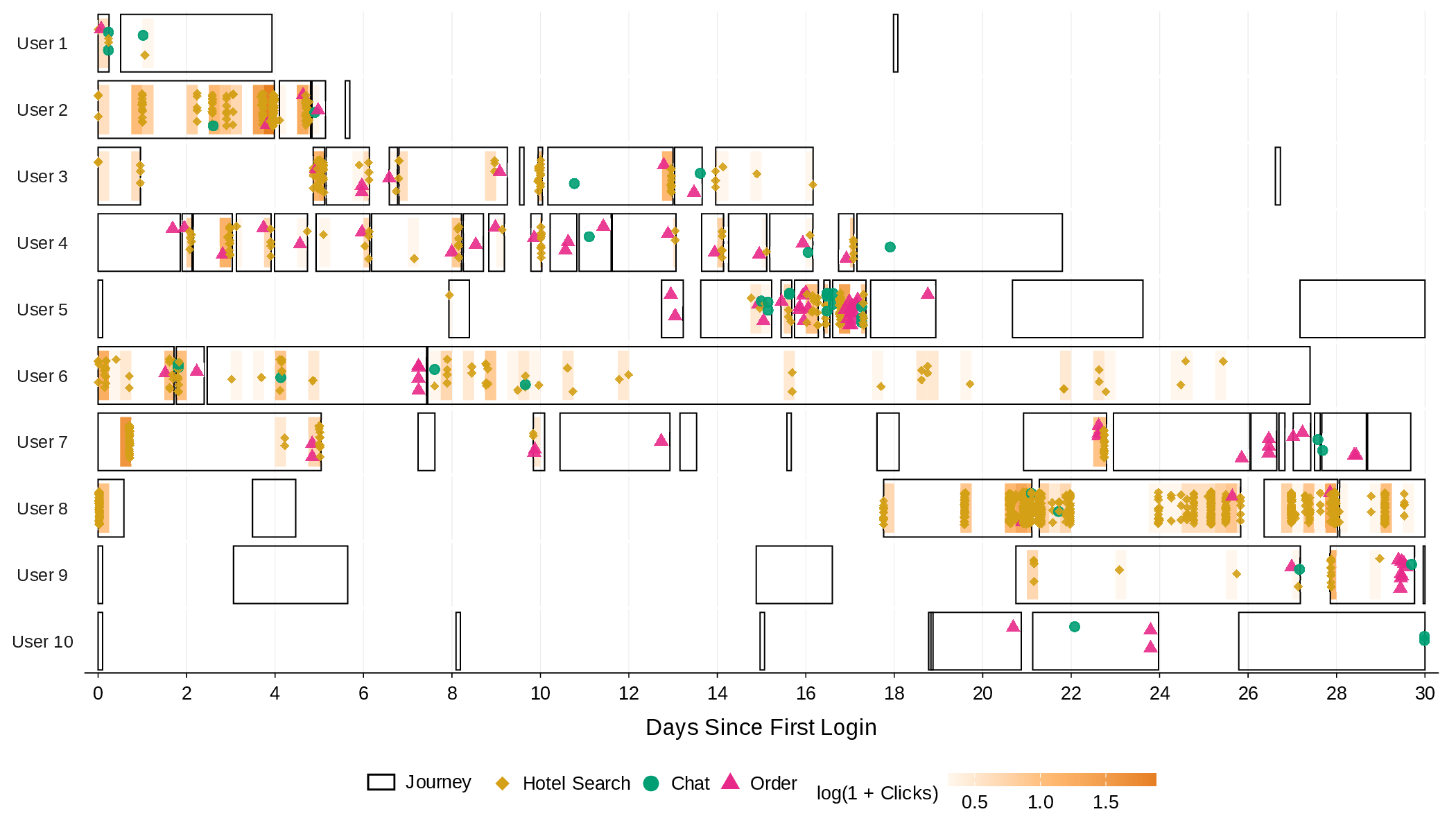}
    \label{fig:journey}
    \begin{minipage}{\linewidth}\setstretch{1}
    {\footnotesize \textit{Notes}: This figure illustrates the journey patterns of ten randomly selected users during the first 30 days after their initial login. Each panel corresponds to one user. The horizontal axis shows the number of days since the user's first login. Key journey events, including hotel searches, chat sessions, orders, and journey endpoints, are marked along each user's timeline.}
    \end{minipage}
\end{figure}

\newpage

\begin{table}[htbp!]
\centering
\caption{Summary statistics of customer journeys}
\label{tab:summary_stats}
\small
\setlength{\tabcolsep}{4pt}
\begin{threeparttable}
\resizebox{\textwidth}{!}{%
\begin{minipage}{\textwidth}
\centering
\renewcommand{\arraystretch}{1.5}
\begin{tabular}{lcccccccc}
\toprule
Variable & N & Mean & S.D. & P5 & P25 & Median & P75 & P95 \\
\midrule
\multicolumn{9}{l}{\textit{\textbf{Panel A: All journeys}}}\\
Journey length (hours)              & 253,209,391 & 34.110  & 79.577  & 0.000 & 0.000  & 7.000  & 43.000 & 138.000     \\
Number of login sessions            & 253,209,391 & 4.235   & 10.717  & 1.000 & 1.000  & 2.000  & 4.000  & 14.000      \\
Total login time (minutes)          & 253,209,391 & 46.951  & 171.683 & 0.000 & 2.000  & 11.000 & 45.000 & 193.000     \\
Number of chat sessions             & 253,209,391 & 0.001   & 0.059   & 0.000 & 0.000  & 0.000  & 0.000  & 0.000       \\
Number of orders                    & 253,209,391 & 0.315   & 0.831   & 0.000 & 0.000  & 0.000  & 0.000  & 2.000       \\
Number of clicks                    & 253,209,391 & 3.747   & 26.450  & 0.000 & 0.000  & 0.000  & 1.000  & 17.000      \\
Number of queries                   & 253,209,391 & 1.885   & 8.748   & 0.000 & 0.000  & 0.000  & 1.000  & 9.000       \\
Number of keyword clicks            & 253,209,391 & 1.085   & 10.133  & 0.000 & 0.000  & 0.000  & 0.000  & 5.000       \\
Number of hotel searches            & 253,209,391 & 8.814   & 101.664 & 0.000 & 0.000  & 0.000  & 2.000  & 34.000      \\
\addlinespace[0.35em]
\multicolumn{9}{l}{\textit{\textbf{Panel B: Journeys with chat vs.\ journeys without chat}}}\\
 & \multicolumn{3}{c}{\textit{With Chat}} & \multicolumn{3}{c}{\textit{Without Chat}} & & \\
\cmidrule(lr){2-4} \cmidrule(lr){5-7}
 & N & Mean & S.D. & N & Mean & S.D. & $t$-stat & $p$-value \\
\cmidrule(lr){2-4} \cmidrule(lr){5-7} \cmidrule(lr){8-8} \cmidrule(lr){9-9}
Journey length (hours)              & 192,402 & 104.104 & 205.604 & 253,016,990 & 34.057  & 79.381  & 149.430 & $<$0.001 \\
Number of login sessions            & 192,402 & 14.212  & 40.454  & 253,016,990 & 4.227   & 10.659  & 108.266 & $<$0.001 \\
Total login time (minutes)          & 192,402 & 252.631 & 940.980 & 253,016,990 & 46.795  & 169.682 & 95.949  & $<$0.001 \\
Number of chat sessions             & 192,402 & 1.447   & 1.582   & 253,016,990 & 0.000   & 0.000   & 401.275 & $<$0.001 \\
Number of orders                    & 192,402 & 0.508   & 6.099   & 253,016,990 & 0.315   & 0.814   & 13.904  & $<$0.001 \\
Number of clicks                    & 192,402 & 24.141  & 136.916 & 253,016,990 & 3.732   & 26.184  & 65.382  & $<$0.001 \\
Number of queries                   & 192,402 & 6.575   & 33.440  & 253,016,990 & 1.881   & 8.701   & 61.561  & $<$0.001 \\
Number of keyword clicks            & 192,402 & 6.625   & 43.227  & 253,016,990 & 1.081   & 10.066  & 56.256  & $<$0.001 \\
Number of hotel searches            & 192,402 & 52.897  & 456.480 & 253,016,990 & 8.780   & 100.913 & 42.391  & $<$0.001 \\
\bottomrule
\end{tabular}
\small
\begin{tablenotes}[flushleft]
\item \textit{Notes}: This table reports journey-level summary statistics. Panel A reports the number of observations, mean, standard deviation, and selected percentiles for all journeys. Panel B compares journeys with at least one AI chat session to journeys without chat. The reported $t$-statistics and $p$-values are from tests of equal means across the two groups. All statistics are rounded to three decimal places.
\end{tablenotes}
\end{minipage}%
}
\end{threeparttable}
\end{table}

\newpage

\begin{figure}[htbp!]
    \caption{Normalized temporal distribution of events relative to chat sessions}
\centering\includegraphics[width=0.9\linewidth]{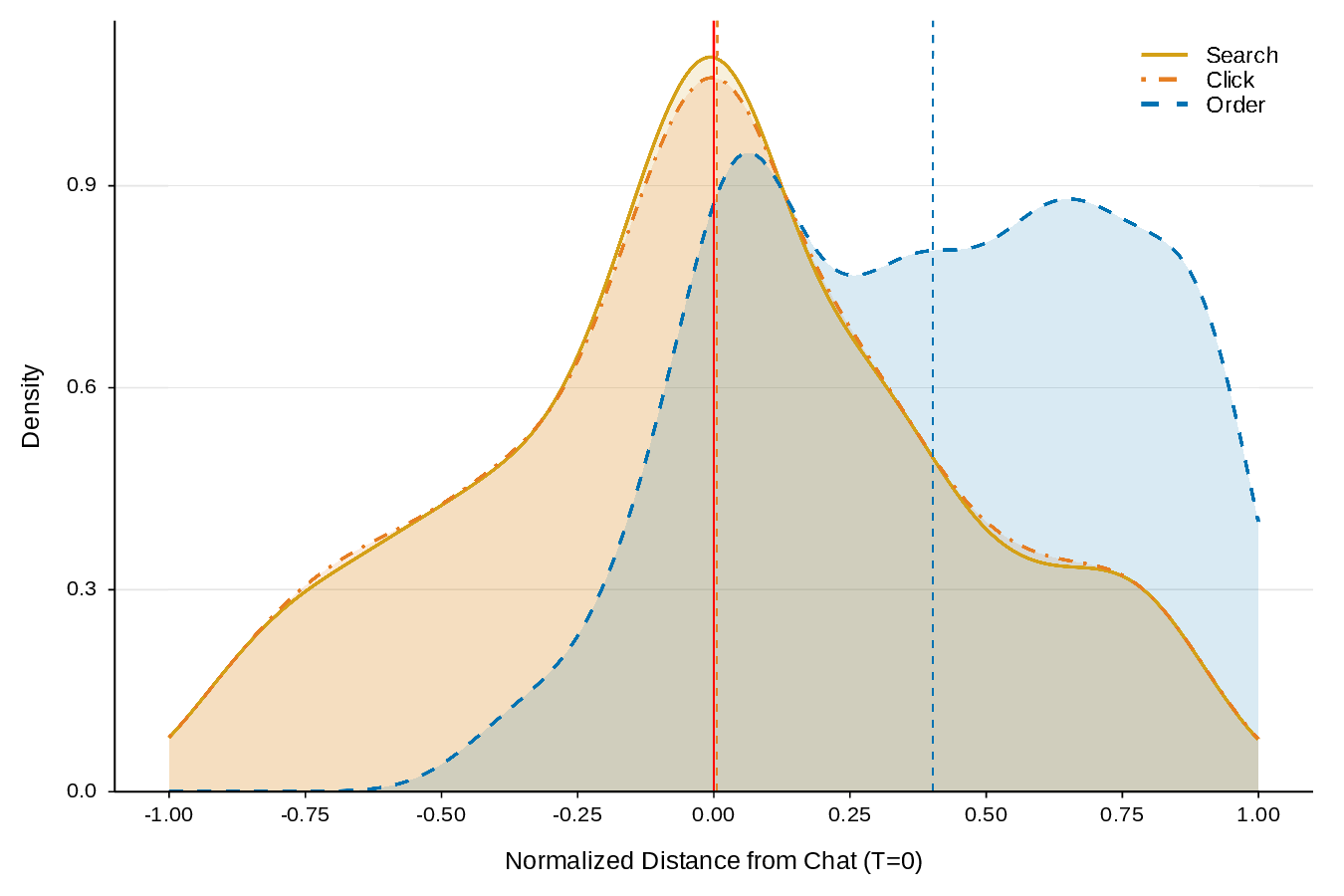}
    \label{fig:norm_chat}
    \begin{minipage}{0.9\linewidth}\setstretch{1}
    {\footnotesize \textit{Notes}: This figure shows the kernel density estimates of the normalized temporal distance between each event type (hotel searches, clicks, and orders) and the chat session, where the chat event is set as the time origin ($T=0$). For each journey, the median timestamp of each event type is computed, and the normalized distance is calculated as (median event time $-$ median chat time) / journey duration, so that negative values indicate events occurring before the chat and positive values indicate events occurring after. Each journey contributes exactly one observation per event type. The sample is restricted to order-journeys containing all four activity types (orders, chats, hotel searches, and clicks). The vertical dashed lines indicate the median of the normalized distance for each event type. The vertical solid red line marks $T=0$ (the chat event).}
    \end{minipage}
\end{figure}

\newpage

\begin{figure}[htbp!]
    \caption{Normalized temporal distribution of events relative to order placement (with-chat journeys)}
\centering\includegraphics[width=0.9\linewidth]{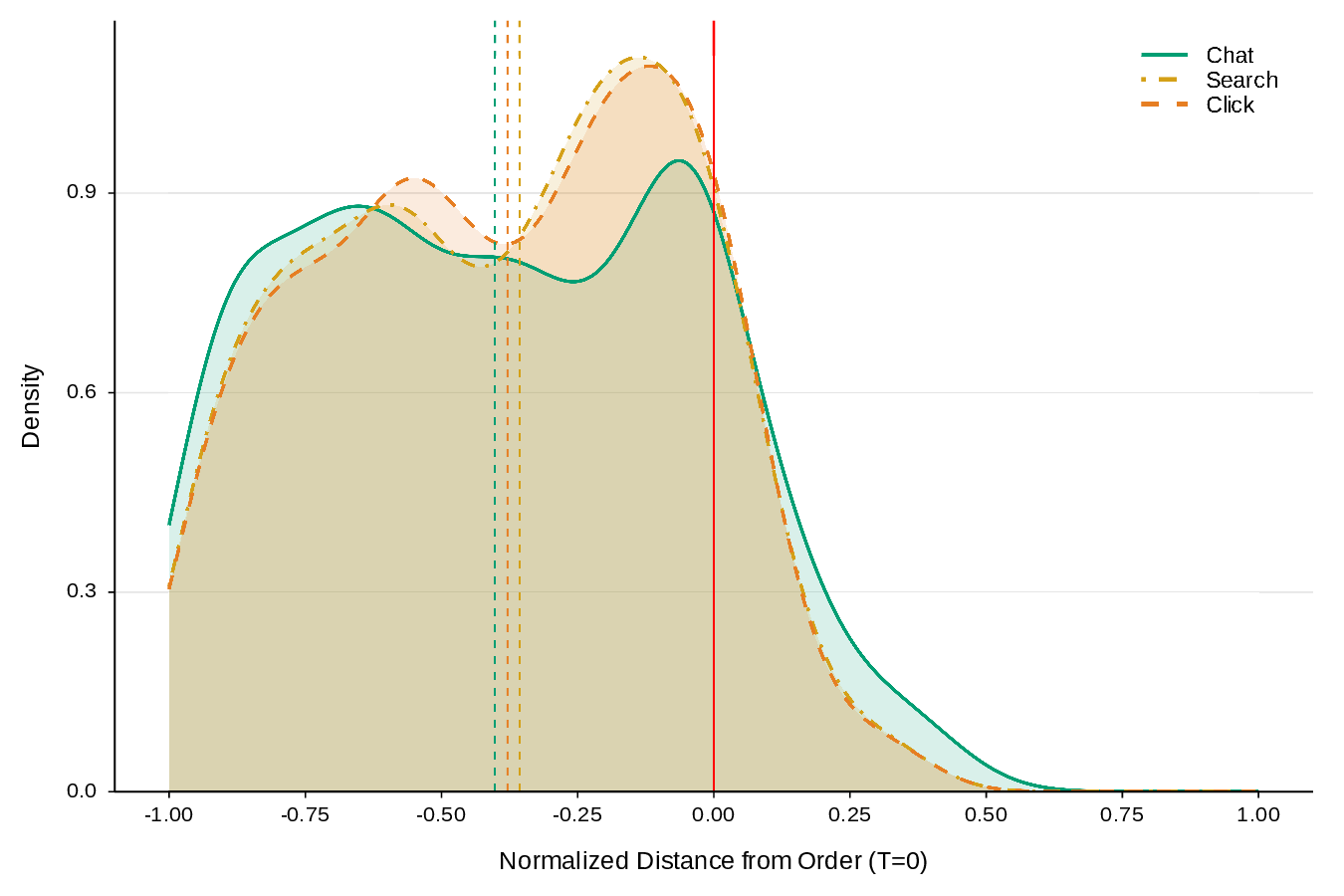}
    \label{fig:norm_order_wc}
    \begin{minipage}{0.9\linewidth}\setstretch{1}
    {\footnotesize \textit{Notes}: This figure shows the kernel density estimates of the normalized temporal distance between each event type (chat sessions, hotel searches, and clicks) and the order placement, where the order event is set as the time origin ($T=0$). For each journey, the median timestamp of each event type is computed, and the normalized distance is calculated as (median event time $-$ median order time) / journey duration, so that negative values indicate events occurring before the order and positive values indicate events occurring after. Each journey contributes exactly one observation per event type. The sample is restricted to order-journeys containing all four activity types (orders, chats, hotel searches, and clicks). The vertical dashed lines indicate the median of the normalized distance for each event type. The vertical solid red line marks $T=0$ (the order event).}
    \end{minipage}
\end{figure}

\newpage

\begin{figure}[htbp!]
    \caption{Normalized temporal distribution of events relative to order placement (no-chat journeys)}
\centering\includegraphics[width=0.9\linewidth]{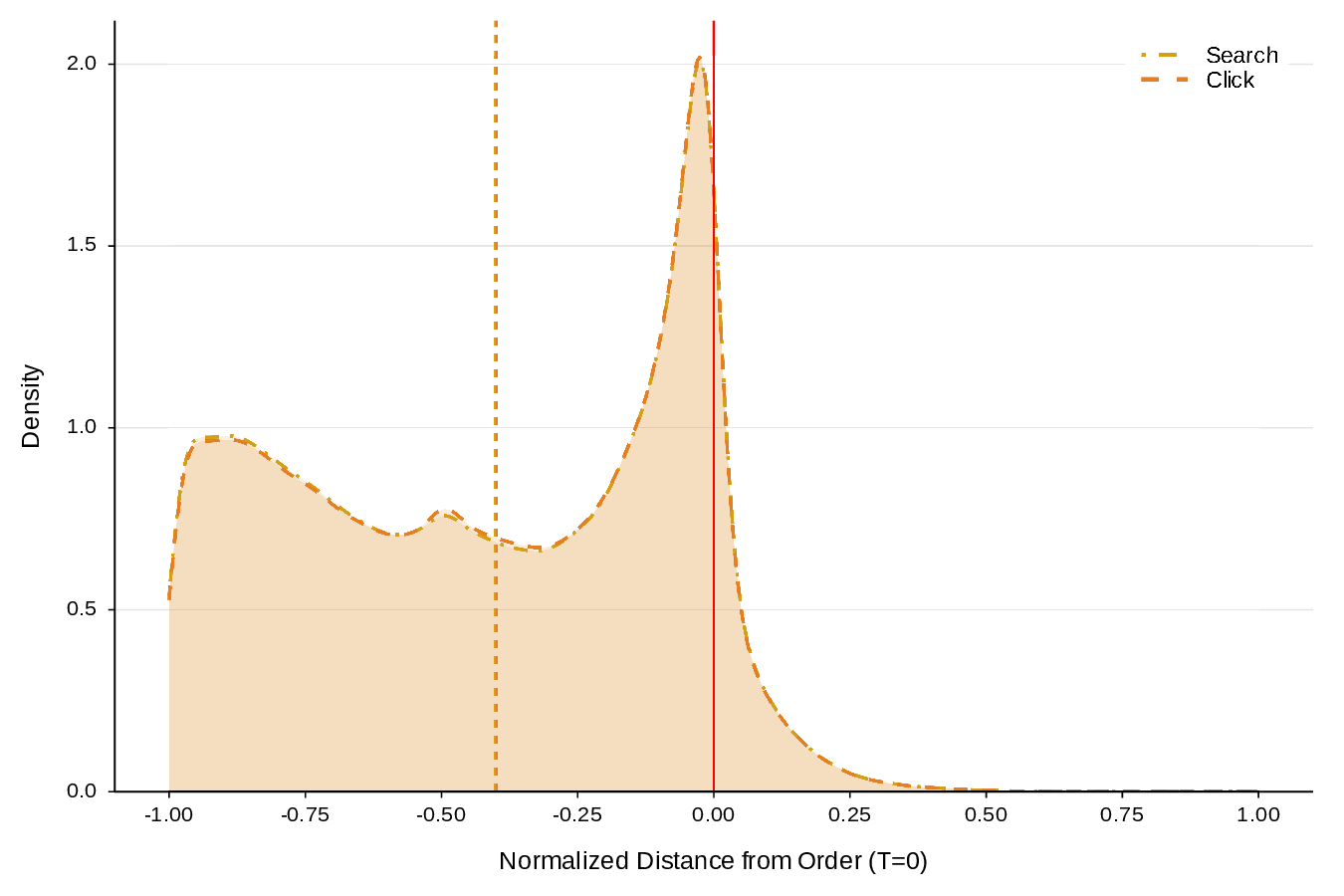}
    \label{fig:norm_order_nc}
    \begin{minipage}{0.9\linewidth}\setstretch{1}
    {\footnotesize \textit{Notes}: This figure shows the kernel density estimates of the normalized temporal distance between each event type (hotel searches and clicks) and the order placement, where the order event is set as the time origin ($T=0$). For each journey, the median timestamp of each event type is computed, and the normalized distance is calculated as (median event time $-$ median order time) / journey duration, so that negative values indicate events occurring before the order and positive values indicate events occurring after. Each journey contributes exactly one observation per event type. The sample is restricted to order-journeys that contain hotel searches and clicks but no chat sessions. The vertical dashed lines indicate the median of the normalized distance for each event type. The vertical solid red line marks $T=0$ (the order event).}
    \end{minipage}
\end{figure}

\newpage


\begin{table}[htbp!]
\centering
\caption{Session Characteristics by Relative Timing of Chat and Search}
\label{tab:chat_sequence_session}
\footnotesize
\begin{threeparttable}
\begin{tabular}{lcccc}
\toprule
 & Chat-Before-Search & Chat-Between-Search & Chat-After-Search & Chat-Only \\
\midrule
\multicolumn{5}{l}{\textit{\textbf{Panel A: Prevalence}}} \\
Share of sessions (\%) & 10.6 & 12.0 & 10.2 & 67.2 \\
\addlinespace[0.35em]
\multicolumn{5}{l}{\textit{\textbf{Panel B: Session characteristics}}} \\
Mean session length (hours) & 0.8 & 1.2 & 0.8 & 0.4 \\
Mean chat sessions & 1.3 & 1.4 & 1.3 & 1.4 \\
Mean hotel searches & 11.3 & 35.8 & 11.2 & 0.0 \\
Mean clicks & 5.1 & 14.6 & 4.8 & 0.0 \\
Mean orders & 0.02 & 0.06 & 0.02 & 0.01 \\
Conversion rate (\%) & 1.1 & 1.9 & 1.1 & 0.4 \\
\addlinespace[0.35em]
\multicolumn{5}{l}{\textit{\textbf{Panel C: Chat intent distribution (\%)}}} \\
Attraction & 38.6 & 37.3 & 43.3 & 43.4 \\
Hotel & 33.2 & 43.1 & 29.9 & 8.8 \\
Travel Planning & 6.4 & 3.3 & 4.4 & 8.6 \\
Customer Service & 4.1 & 3.1 & 4.1 & 6.3 \\
Flight & 0.6 & 0.2 & 0.5 & 2.5 \\
Other Transportation & 4.2 & 3.5 & 4.6 & 10.4 \\
Other & 12.9 & 9.6 & 13.2 & 20.0 \\
\bottomrule
\end{tabular}
\begin{tablenotes}[flushleft]
\item \textit{Notes}:  This table summarizes sessions containing at least one AI chat event. \textit{Chat-Before-Search}, \textit{Chat-Between-Search}, and \textit{Chat-After-Search} sessions each contain at least one hotel search, with the classification based on the relative ordering of chat and search events within the session. \textit{Chat-Before-Search} sessions have chat events occurring before search events, \textit{Chat-After-Search} sessions have chat events occurring after search events, and \textit{Chat-Between-Search} sessions have chat and search events interleaved. \textit{Chat-Only} sessions contain at least one chat event but no hotel search. Panel A reports the share of sessions in each group. Panel B reports mean session-level characteristics. Conversion rate is the percentage of sessions with at least one order. Panel C reports the distribution of chat intent categories within each group; intent shares sum to 100\% within each column, up to rounding. The sample contains 206,245 sessions.
\end{tablenotes}
\end{threeparttable}
\end{table}


\begin{table}[htbp!]
\centering
\caption{Journey Characteristics by Relative Timing of Chat and Search (48-Hour Inactivity Definition)}
\label{tab:chat_sequence_48h}
\footnotesize
\begin{threeparttable}
\begin{tabular}{lcccc}
\toprule
 & Chat-Before-Search & Chat-Between-Search & Chat-After-Search & Chat-Only \\
\midrule
\multicolumn{5}{l}{\textit{\textbf{Panel A: Prevalence}}} \\
Share of journeys (\%) & 15.049 & 30.804 & 12.411 & 41.736 \\
\addlinespace[0.35em]
\multicolumn{5}{l}{\textit{\textbf{Panel B: Journey characteristics}}} \\
Median journey length (hours) & 51.727 & 94.904 & 51.889 & 18.160 \\
Mean login sessions & 11.282 & 28.639 & 10.521 & 5.562 \\
Mean chat sessions & 1.377 & 1.565 & 1.357 & 1.411 \\
Mean hotel searches & 24.196 & 149.860 & 21.350 & 0.000 \\
Mean clicks & 11.423 & 68.307 & 9.468 & 0.000 \\
Mean orders & 0.549 & 0.646 & 0.451 & 0.222 \\
Conversion rate (\%) & 39.292 & 42.977 & 31.153 & 18.343 \\
\addlinespace[0.35em]
\multicolumn{5}{l}{\textit{\textbf{Panel C: Chat intent distribution (\%)}}} \\
Attraction & 44.208 & 41.818 & 48.406 & 39.955 \\
Hotel & 17.965 & 30.469 & 17.690 & 7.056 \\
Travel Planning & 9.832 & 6.081 & 5.399 & 8.021 \\
Customer Service & 4.874 & 3.746 & 5.566 & 6.926 \\
Flight & 1.372 & 0.587 & 0.925 & 3.233 \\
Other Transportation & 6.231 & 4.854 & 5.649 & 12.668 \\
Other & 15.518 & 12.445 & 16.364 & 22.141 \\
\bottomrule
\end{tabular}
\begin{tablenotes}[flushleft]
\item \textit{Notes}: This table reports a robustness version of Table~\ref{tab:chat_sequence} using an alternative journey definition based only on a 48-hour inactivity rule. Consecutive login sessions are grouped into the same journey unless the gap between one session's logout and the next session's login is at least 48 hours; unlike the baseline definition, no additional order-based splitting rule is applied. The table compares four groups of journeys containing at least one AI chat event. The first three groups also contain at least one hotel search and are classified by the relative ordering of chat and search within the journey: \textit{Chat-Before-Search}, \textit{Chat-Between-Search}, and \textit{Chat-After-Search}. \textit{Chat-Only} journeys contain chat but no hotel search. Panel A reports the share of journeys in each group. Panel B reports journey-level characteristics; conversion rate denotes the share of journeys with at least one order. Panel C reports the distribution of chat intent categories within each group; shares sum to 100\% within each column, up to rounding. Differences across the three chat-and-search groups are statistically significant at the 0.1\% level, and the chat-intent distribution also differs significantly across these groups ($\chi^2 = 4{,}419.1$, $p < 0.001$).
\end{tablenotes}
\end{threeparttable}
\end{table}


\begin{table}[htbp!]
\centering
\caption{Journey Characteristics by Relative Timing of Chat and Search (48-Hour Inactivity, 8-Hour Order-Gap Definition)}
\label{tab:chat_sequence_48h_8h}
\footnotesize
\begin{threeparttable}
\begin{tabular}{lcccc}
\toprule
 & Chat-Before-Search & Chat-Between-Search & Chat-After-Search & Chat-Only \\
\midrule
\multicolumn{5}{l}{\textit{\textbf{Panel A: Prevalence}}} \\
Share of journeys (\%) & 15.089 & 30.940 & 12.414 & 41.557 \\
\addlinespace[0.35em]
\multicolumn{5}{l}{\textit{\textbf{Panel B: Journey characteristics}}} \\
Median journey length (hours) & 51.878 & 94.874 & 52.037 & 18.472 \\
Mean login sessions & 11.319 & 28.620 & 10.548 & 5.586 \\
Mean chat sessions & 1.377 & 1.566 & 1.358 & 1.411 \\
Mean hotel searches & 24.397 & 150.109 & 21.415 & 0.000 \\
Mean clicks & 11.524 & 68.420 & 9.528 & 0.000 \\
Mean orders & 0.571 & 0.705 & 0.473 & 0.225 \\
Conversion rate (\%) & 39.430 & 43.206 & 31.137 & 17.957 \\
\addlinespace[0.35em]
\multicolumn{5}{l}{\textit{\textbf{Panel C: Chat intent distribution (\%)}}} \\
Attraction & 44.217 & 41.884 & 48.492 & 39.861 \\
Hotel & 17.902 & 30.374 & 17.681 & 7.063 \\
Travel Planning & 9.840 & 6.071 & 5.387 & 8.035 \\
Customer Service & 4.879 & 3.770 & 5.532 & 6.928 \\
Flight & 1.370 & 0.589 & 0.922 & 3.244 \\
Other Transportation & 6.239 & 4.881 & 5.600 & 12.693 \\
Other & 15.553 & 12.430 & 16.386 & 22.177 \\
\bottomrule
\end{tabular}
\begin{tablenotes}[flushleft]
\item \textit{Notes}: This table reports a robustness version of Table~\ref{tab:chat_sequence} using an alternative journey definition that combines adjacent login sessions into the same journey unless the inactivity gap is at least 48 hours, and additionally splits journeys between adjacent order-containing sessions when the inactivity gap exceeds 8 hours. The table compares four groups of journeys containing at least one AI chat event. The first three groups also contain at least one hotel search and are classified by the relative ordering of chat and search within the journey: \textit{Chat-Before-Search}, \textit{Chat-Between-Search}, and \textit{Chat-After-Search}. \textit{Chat-Only} journeys contain chat but no hotel search. Panel A reports the share of journeys in each group. Panel B reports journey-level characteristics; conversion rate denotes the share of journeys with at least one order. Panel C reports the distribution of chat intent categories within each group; shares sum to 100\% within each column, up to rounding. Differences across the three chat-and-search groups are statistically significant at the 0.1\% level, and the chat-intent distribution also differs significantly across these groups ($\chi^2 = 4{,}411.5$, $p < 0.001$).
\end{tablenotes}
\end{threeparttable}
\end{table}


\begin{table}[htbp!]
\centering
\caption{Journey Characteristics by Relative Timing of Chat and Search (48-Hour Inactivity, 12-Hour Order-Gap Definition)}
\label{tab:chat_sequence_48h_12h}
\footnotesize
\begin{threeparttable}
\begin{tabular}{lcccc}
\toprule
 & Chat-Before-Search & Chat-Between-Search & Chat-After-Search & Chat-Only \\
\midrule
\multicolumn{5}{l}{\textit{\textbf{Panel A: Prevalence}}} \\
Share of journeys (\%) & 15.091 & 30.957 & 12.420 & 41.531 \\
\addlinespace[0.35em]
\multicolumn{5}{l}{\textit{\textbf{Panel B: Journey characteristics}}} \\
Median journey length (hours) & 51.938 & 94.876 & 52.049 & 18.526 \\
Mean login sessions & 11.324 & 28.617 & 10.553 & 5.590 \\
Mean chat sessions & 1.377 & 1.566 & 1.358 & 1.411 \\
Mean hotel searches & 24.428 & 150.159 & 21.437 & 0.000 \\
Mean clicks & 11.537 & 68.441 & 9.534 & 0.000 \\
Mean orders & 0.578 & 0.716 & 0.481 & 0.226 \\
Conversion rate (\%) & 39.432 & 43.232 & 31.170 & 17.898 \\
\addlinespace[0.35em]
\multicolumn{5}{l}{\textit{\textbf{Panel C: Chat intent distribution (\%)}}} \\
Attraction & 44.230 & 41.881 & 48.515 & 39.849 \\
Hotel & 17.878 & 30.367 & 17.664 & 7.066 \\
Travel Planning & 9.847 & 6.069 & 5.386 & 8.036 \\
Customer Service & 4.888 & 3.772 & 5.531 & 6.925 \\
Flight & 1.370 & 0.591 & 0.915 & 3.247 \\
Other Transportation & 6.241 & 4.888 & 5.571 & 12.700 \\
Other & 15.545 & 12.431 & 16.419 & 22.176 \\
\bottomrule
\end{tabular}
\begin{tablenotes}[flushleft]
\item \textit{Notes}: This table reports a robustness version of Table~\ref{tab:chat_sequence} using an alternative journey definition that combines adjacent login sessions into the same journey unless the inactivity gap is at least 48 hours, and additionally splits journeys between adjacent order-containing sessions when the inactivity gap exceeds 12 hours. The table compares four groups of journeys containing at least one AI chat event. The first three groups also contain at least one hotel search and are classified by the relative ordering of chat and search within the journey: \textit{Chat-Before-Search}, \textit{Chat-Between-Search}, and \textit{Chat-After-Search}. \textit{Chat-Only} journeys contain chat but no hotel search. Panel A reports the share of journeys in each group. Panel B reports journey-level characteristics; conversion rate denotes the share of journeys with at least one order. Panel C reports the distribution of chat intent categories within each group; shares sum to 100\% within each column, up to rounding. Differences across the three chat-and-search groups are statistically significant at the 0.1\% level, and the chat-intent distribution also differs significantly across these groups ($\chi^2 = 4{,}425.3$, $p < 0.001$).
\end{tablenotes}
\end{threeparttable}
\end{table}

\newpage


\begin{figure}[htbp!]
    \caption{Density of Event Types over the Course of the Customer Journey (48-Hour Inactivity Definition)}
\centering\includegraphics[width=0.9\linewidth]{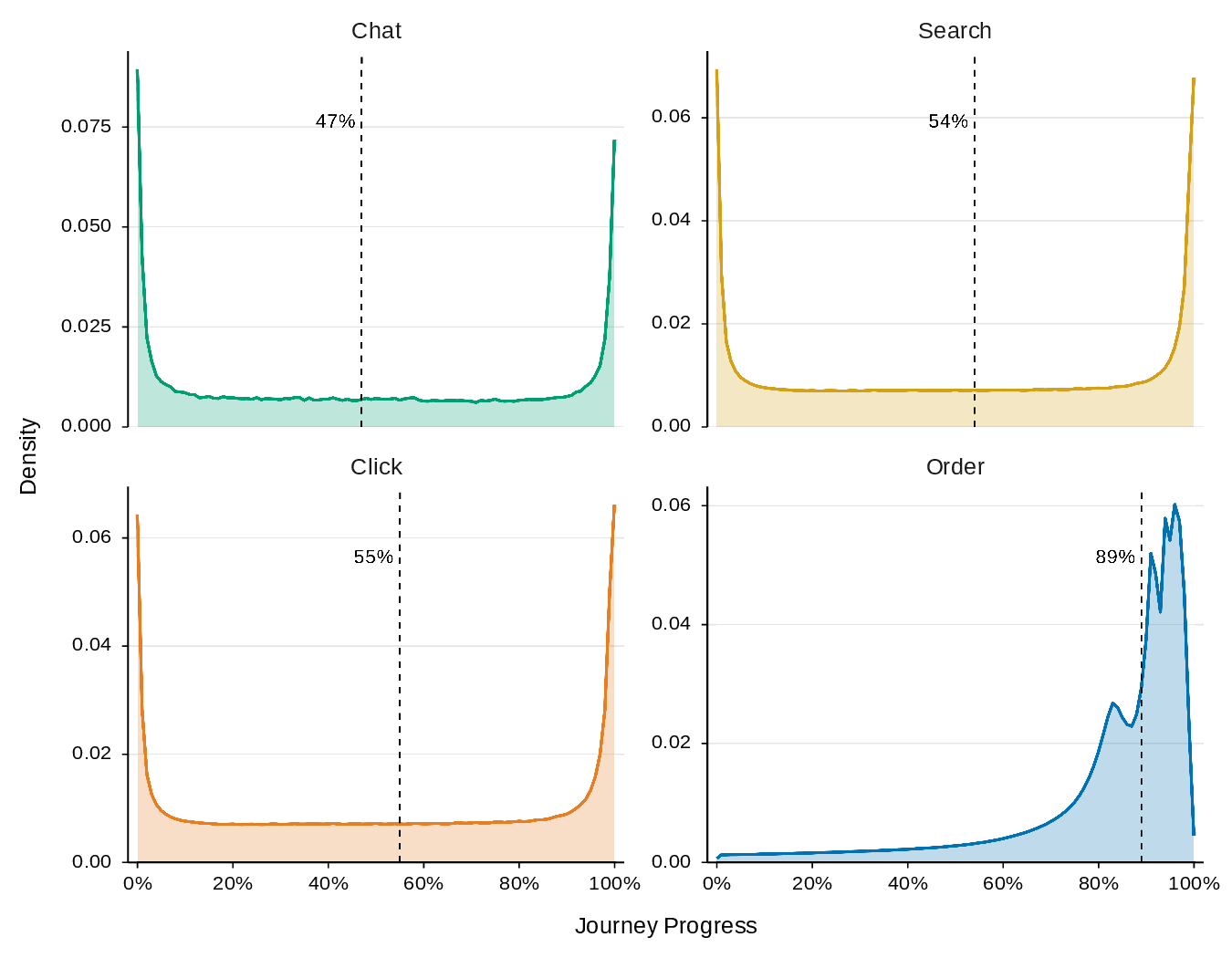}
    \label{fig:progression_density_48h}
    \begin{minipage}{0.9\linewidth}\setstretch{1}
    {\footnotesize \textit{Notes}: This figure reports a robustness check of Figure~\ref{fig:progression_density} using an alternative journey definition that combines adjacent login sessions into a single journey unless the inactivity gap between one session's logout and the next session's login is at least 48 hours, without applying the additional order-based splitting rule used in the baseline definition. Each panel displays the kernel density estimate of one of four event types---clicks, orders, chats, and hotel searches---over the normalized progression of the customer journey from 0\% to 100\%. The horizontal axis represents journey progress as a percentage of total journey duration, and the vertical axis represents density. The vertical dashed line in each panel indicates the median journey progress at which the corresponding event occurs.
    }
    \end{minipage}
\end{figure}

\newpage

\newpage


\begin{figure}[htbp!]
    \caption{Density of Event Types over the Course of the Customer Journey (48-Hour Inactivity, 8-Hour Order-Gap Definition)}
\centering\includegraphics[width=0.9\linewidth]{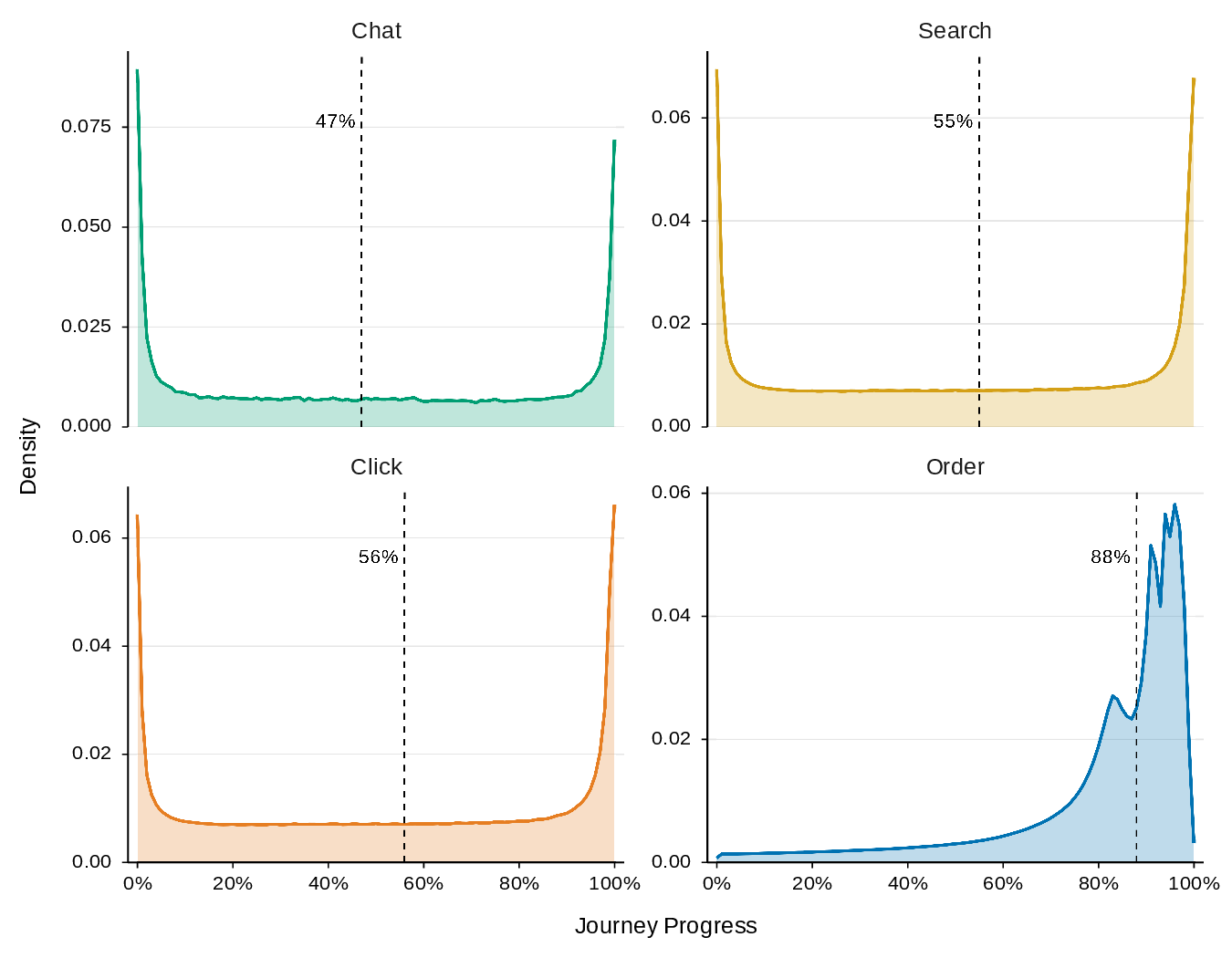}
    \label{fig:progression_density_48h_8h}
    \begin{minipage}{0.9\linewidth}\setstretch{1}
    {\footnotesize \textit{Notes}: This figure reports a robustness check of Figure~\ref{fig:progression_density} using an alternative journey definition that combines adjacent login sessions into a single journey unless the inactivity gap is at least 48 hours, and splits journeys between adjacent order-containing sessions whenever the inactivity gap exceeds 8 hours rather than the 4-hour threshold used in the baseline definition. Each panel displays the kernel density estimate of one of four event types---clicks, orders, chats, and hotel searches---over the normalized progression of the customer journey from 0\% to 100\%. The horizontal axis represents journey progress as a percentage of total journey duration, and the vertical axis represents density. The vertical dashed line in each panel indicates the median journey progress at which the corresponding event occurs.
    }
    \end{minipage}
\end{figure}

\newpage


\begin{figure}[htbp!]
    \caption{Density of Event Types over the Course of the Customer Journey (48-Hour Inactivity, 12-Hour Order-Gap Definition)}
\centering\includegraphics[width=0.9\linewidth]{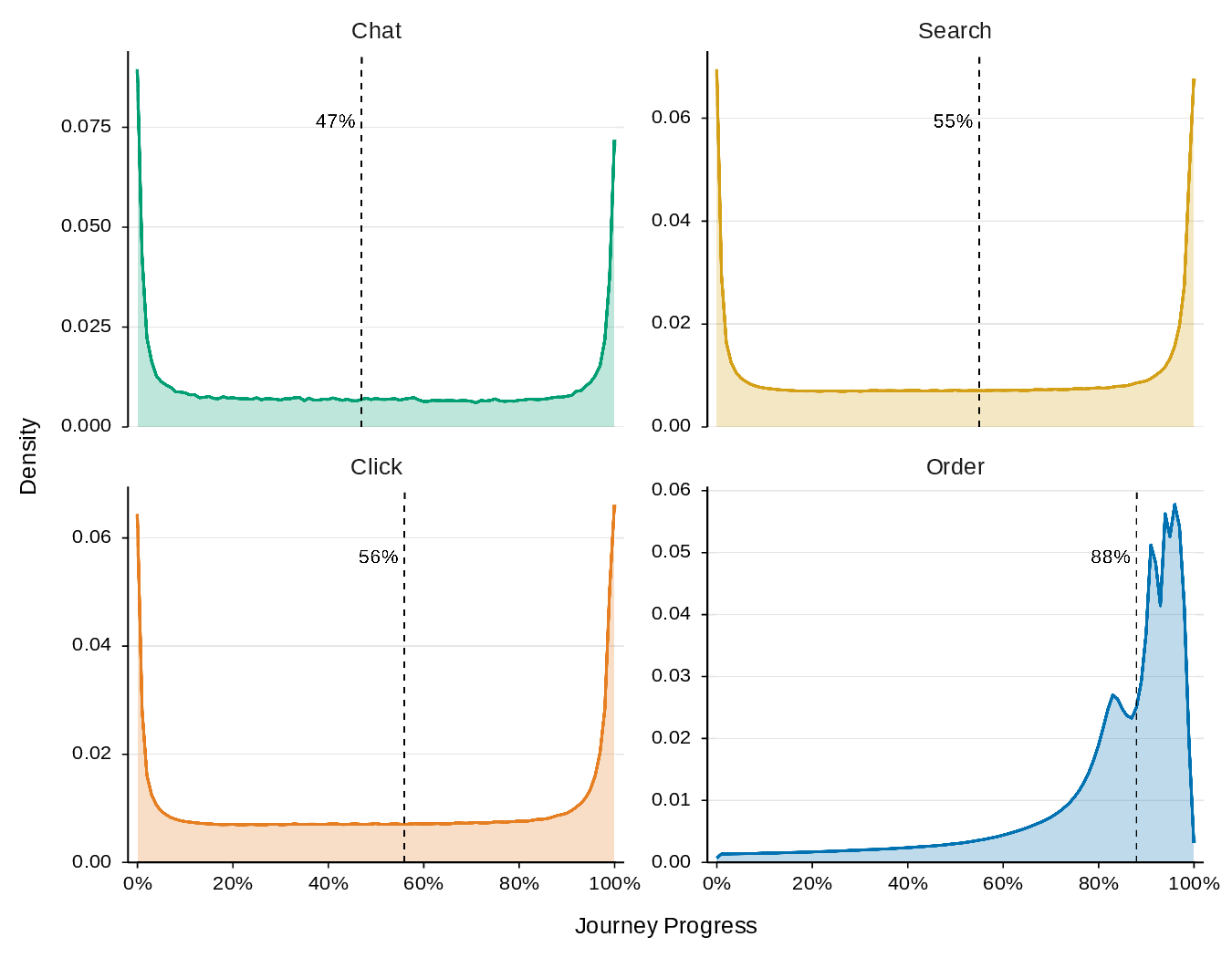}
    \label{fig:progression_density_48h_12h}
    \begin{minipage}{0.9\linewidth}\setstretch{1}
    {\footnotesize \textit{Notes}: This figure reports a robustness check of Figure~\ref{fig:progression_density} using an alternative journey definition that combines adjacent login sessions into a single journey unless the inactivity gap is at least 48 hours, and splits journeys between adjacent order-containing sessions whenever the inactivity gap exceeds 12 hours rather than the 4-hour threshold used in the baseline definition. Each panel displays the kernel density estimate of one of four event types---clicks, orders, chats, and hotel searches---over the normalized progression of the customer journey from 0\% to 100\%. The horizontal axis represents journey progress as a percentage of total journey duration, and the vertical axis represents density. The vertical dashed line in each panel indicates the median journey progress at which the corresponding event occurs.
    }
    \end{minipage}
\end{figure}

\newpage

\begin{table}[htbp!]
\centering
\caption{Illustrative AI Chat Conversations by Intent Category}
\label{tab:example_conversations}
\begin{threeparttable}
\scriptsize
\begin{tabular}{p{3cm}p{4.5cm}p{6.5cm}}
\toprule
Intent & User Query (Translated) & AI Response (Excerpt, Translated) \\
\midrule
Attraction &
What are the must-visit attractions in Qingdao? &
Highlights include Zhanqiao Pier, Badaguan Scenic Area, and Laoshan Scenic Area (trails from ¥120). \\
\midrule
Hotel &
Good-value hotels or guesthouses near a subway station in Beijing? &
Six metro-adjacent options are compared, including a historic siheyuan courtyard hotel (255m from station, ¥394/night) and a loft apartment (¥381/night, 4.7 rating). \\
\midrule
Travel Planning &
What is the best one-day itinerary for Macau? &
Start at Macau Tower, continue to the Venetian and Parisian resorts, and end with the figure-eight Ferris wheel at night. \\
\midrule
Consumer Support &
How do I register for Phoenix membership through the Trip.com app? &
Please contact Trip.com customer service via in-app live chat or phone support using the link below. \\
\midrule
Flight &
Which economy-class flights departing Dali have available seats today? &
Which destination city are you flying to? \\
\midrule
Other Transportation &
What is the driving route from Heihe to Wudalianchi? &
The drive is approximately 247 km and takes around 3 hours; see the link below for the full route. \\
\midrule
Other &
Can you share some financial fraud prevention tips? &
This assistant covers travel topics only; please consult a dedicated resource for financial queries. \\
\bottomrule
\end{tabular}
\begin{tablenotes}[flushleft]
\footnotesize
\item \textit{Notes}: This table presents one illustrative AI chat conversation for each intent category. User queries and AI responses are translated from Chinese. The reported responses are brief excerpts rather than full transcripts and are shown for illustration only.
\end{tablenotes}
\end{threeparttable}
\end{table}

\newpage

\begin{table}[htbp!]
\centering
\caption{Summary statistics for AI assistant engagement and usage patterns}
\label{tab:sumstat_ai_assistant}
\small
\setlength{\tabcolsep}{4pt}
\begin{threeparttable}
\resizebox{\textwidth}{!}{%
\begin{minipage}{\textwidth}
\centering
\renewcommand{\arraystretch}{1.5}
\begin{tabular}{lcccccccc}
\toprule
Variable & N & Mean & S.D. & P5 & P25 & Median & P75 & P95 \\
\midrule
\multicolumn{9}{l}{\textit{\textbf{Panel A: User-level AI assistant engagement}}}\\
Total number of logins                              & 31,129,629 & 34.339 & 46.403 & 2.000 & 7.000 & 19.000 & 44.000 & 114.000 \\
Average chat sessions per login among AI adopters                   & 188,647 & 0.092 & 0.171 & 0.006 & 0.016 & 0.035 & 0.087 & 0.333 \\
\addlinespace[0.35em]
\multicolumn{9}{l}{\textit{\textbf{Panel B: Session-level summary statistics}}}\\
Number of requests in session                  & 213,460    & 1.313  & 0.936  & 1.000 & 1.000 & 1.000  & 1.000  & 3.000 \\
Characters per minute                         & 213,460    & 1,456.520 & 3,710.211 & 18.000 & 458.000 & 1,168.079 & 1,656.000 & 2,598.000 \\
Deep-thinking mode used                       & 213,460    & 0.787  & 0.409  & 0.000 & 1.000 & 1.000  & 1.000  & 1.000 \\
\addlinespace[0.35em]
\multicolumn{9}{l}{\textit{\textbf{Panel C: Request-level summary statistics}}}\\
User message length (chars)                   & 280,350    & 10.519 & 13.670 & 2.000 & 5.000 & 8.000  & 12.000 & 25.000 \\
Assistant reply length (chars)                & 280,350    & 1,089.130 & 717.437 & 22.000 & 459.000 & 1,145.000 & 1,596.000 & 2,187.000 \\
User message information entropy              & 280,350    & 1.049  & 0.654  & 0.000 & 1.000 & 1.000  & 1.000  & 2.000 \\
Assistant reply information entropy           & 280,350    & 3.041  & 1.793  & 0.000 & 2.000 & 4.000  & 4.000  & 5.000 \\
Reply flagged as incomplete                   & 280,350    & 0.089  & 0.285  & 0.000 & 0.000 & 0.000  & 0.000  & 1.000 \\
\bottomrule
\end{tabular}
\small
\begin{tablenotes}[flushleft]
\item \textit{Notes}: This table reports summary statistics for AI assistant engagement and usage patterns at three levels of observation. Panel A reports user-level measures, including overall platform login activity and AI assistant usage among adopters. Panel B reports session-level statistics for AI assistant sessions. Panel C reports request-level characteristics for individual user requests and assistant replies. $N$ denotes the number of observations, S.D.\ denotes the standard deviation, and P5, P25, P75, and P95 denote the 5th, 25th, 75th, and 95th percentiles, respectively. For indicator variables, the mean can be interpreted as a proportion. Sample sizes differ across panels because the unit of observation varies. All statistics are rounded to three decimal places.
\end{tablenotes}
\end{minipage}%
}
\end{threeparttable}
\end{table}

\newpage

\begin{table}[!htbp]
\centering
\caption{User intent distribution of AI assistant requests}
\label{tab:intent_share_by_group}
\small
\setlength{\tabcolsep}{4pt}
\begin{threeparttable}
\resizebox{\textwidth}{!}{%
\begin{minipage}{\textwidth}
\centering
\renewcommand{\arraystretch}{1.25}
\begin{tabular}{p{0.26\textwidth}*{7}{p{0.106\textwidth}}}
\toprule
& \multicolumn{7}{c}{User Intent} \\
\cmidrule(lr){2-8}
User group & Attr. & Hotel & Plan. & C.S. & Flight & Trans. & Other \\
\midrule
\multicolumn{8}{l}{\textit{\textbf{Panel A: Full sample}}}\\
All users                    & 0.421 & 0.177 & 0.073 & 0.054 & 0.018 & 0.083 & 0.173 \\
\addlinespace[0.35em]
\multicolumn{8}{l}{\textit{\textbf{Panel B: Gender}}}\\
Female                       & 0.450 & 0.175 & 0.068 & 0.053 & 0.015 & 0.076 & 0.163 \\
Male                         & 0.408 & 0.171 & 0.083 & 0.056 & 0.020 & 0.079 & 0.183 \\
\addlinespace[0.35em]
\multicolumn{8}{l}{\textit{\textbf{Panel C: Age group}}}\\
Age 24 and below             & 0.352 & 0.148 & 0.057 & 0.113 & 0.021 & 0.099 & 0.209 \\
Age 25--34                   & 0.412 & 0.175 & 0.079 & 0.079 & 0.018 & 0.052 & 0.184 \\
Age 35--49                   & 0.478 & 0.162 & 0.082 & 0.044 & 0.013 & 0.069 & 0.152 \\
Age 50+                      & 0.377 & 0.211 & 0.059 & 0.026 & 0.022 & 0.117 & 0.188 \\
\addlinespace[0.35em]
\multicolumn{8}{l}{\textit{\textbf{Panel D: City level}}}\\
Municipality (highest)       & 0.443 & 0.182 & 0.079 & 0.051 & 0.015 & 0.055 & 0.174 \\
Provincial capital           & 0.425 & 0.181 & 0.075 & 0.057 & 0.017 & 0.073 & 0.172 \\
Prefecture city              & 0.412 & 0.171 & 0.071 & 0.053 & 0.020 & 0.102 & 0.172 \\
County-level city (lowest)   & 0.413 & 0.175 & 0.062 & 0.052 & 0.020 & 0.108 & 0.171 \\
Intl/SAR/Other               & 0.352 & 0.166 & 0.056 & 0.072 & 0.041 & 0.065 & 0.249 \\
\addlinespace[0.35em]
\multicolumn{8}{l}{\textit{\textbf{Panel E: VIP level}}}\\
Silver VIP                   & 0.377 & 0.159 & 0.067 & 0.060 & 0.021 & 0.126 & 0.190 \\
Gold VIP                     & 0.445 & 0.188 & 0.077 & 0.048 & 0.015 & 0.066 & 0.161 \\
Platinum VIP                 & 0.481 & 0.173 & 0.078 & 0.053 & 0.013 & 0.046 & 0.156 \\
Diamond VIP                  & 0.479 & 0.166 & 0.063 & 0.065 & 0.013 & 0.048 & 0.165 \\
Golden Diamond VIP           & 0.459 & 0.154 & 0.059 & 0.074 & 0.012 & 0.033 & 0.208 \\
Black Diamond VIP            & 0.469 & 0.149 & 0.030 & 0.100 & 0.010 & 0.033 & 0.209 \\
\addlinespace[0.35em]
\multicolumn{8}{l}{\textit{\textbf{Panel F: Device}}}\\
iOS                          & 0.447 & 0.178 & 0.060 & 0.068 & 0.016 & 0.053 & 0.178 \\
Android                      & 0.408 & 0.176 & 0.080 & 0.047 & 0.019 & 0.099 & 0.171 \\
\addlinespace[0.35em]
\multicolumn{8}{l}{\textit{\textbf{Panel G: New vs.\ existing users}}}\\
New user                     & 0.400 & 0.174 & 0.072 & 0.061 & 0.019 & 0.092 & 0.181 \\
Existing user                & 0.445 & 0.179 & 0.075 & 0.046 & 0.017 & 0.073 & 0.165 \\
\bottomrule
\end{tabular}
\small
\begin{tablenotes}[flushleft]
\item \textit{Notes}: This table reports the distribution of AI assistant request intents across user groups. Each entry is the share of requests in the corresponding user group classified into a given intent category. Shares sum to one within each row, up to rounding. Column labels are abbreviated as follows: Attr.\ = Attraction, Plan.\ = Travel Planning, C.S.\ = Customer Service, and Trans.\ = Other Transportation.
\end{tablenotes}
\end{minipage}%
}
\end{threeparttable}
\end{table}

\newpage

\begin{figure}[H]
\centering
    \caption{Average number of AI chat sessions per user by demographic and account characteristics}
    \label{fig:avg_sessions_per_user}
\includegraphics[width=0.8\linewidth]{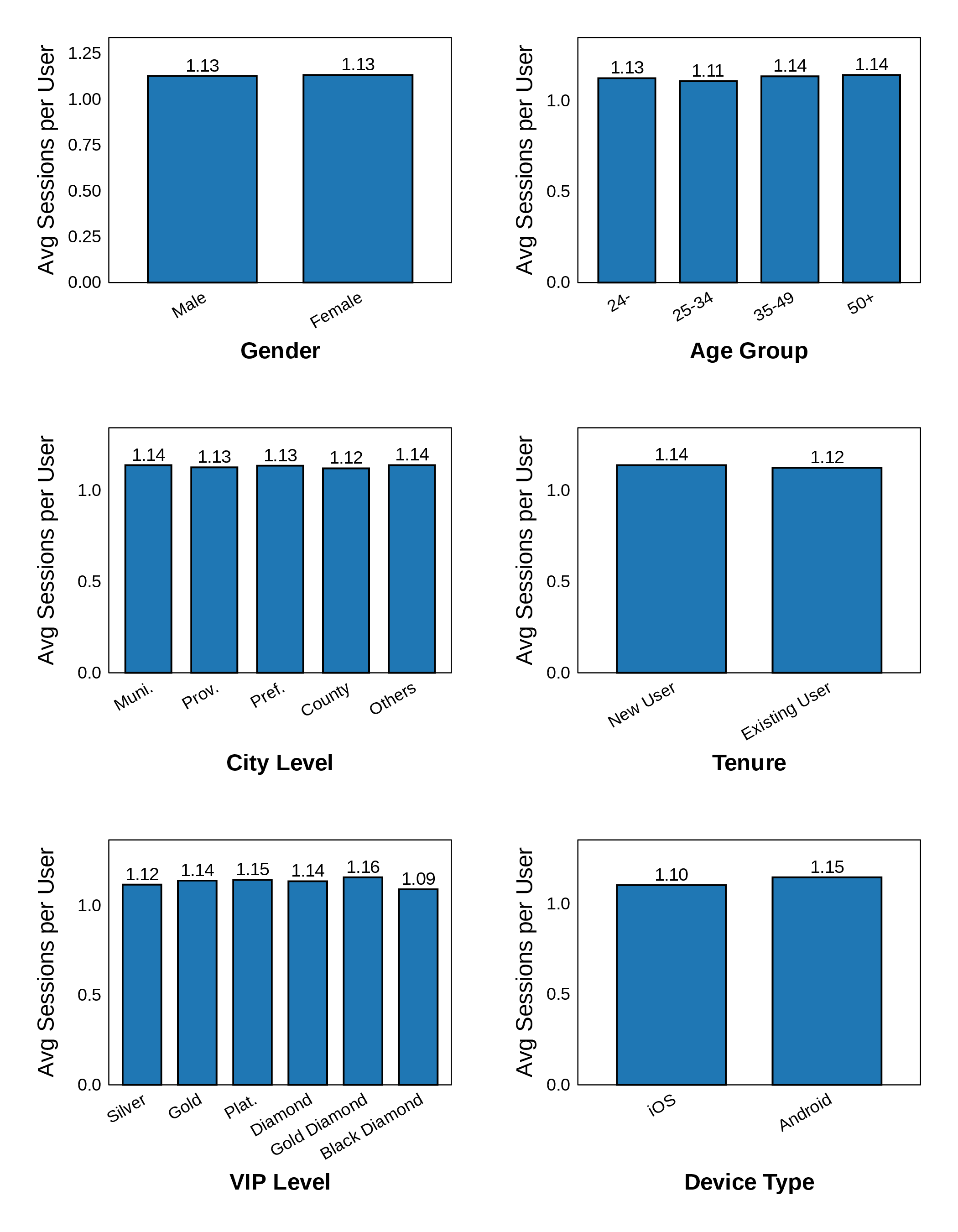}
    \begin{minipage}{0.8\linewidth}
    \setstretch{1}
    {\footnotesize \textit{Notes}: This figure shows the average number of AI chat sessions per user across six user characteristics: gender, age group, city level, tenure (new versus existing users), VIP level (Silver, Gold, Platinum, Diamond, Gold Diamond, and Black Diamond), and device type (Android versus iOS). Each panel reports the mean number of chat sessions for the categories within the corresponding characteristic. The averages are computed among users with at least one AI chat session during the sample period.}
    \end{minipage}
\end{figure}

\newpage

\begin{figure}[H]
\centering
    \caption{Average AI chat session duration by demographic and account characteristics}
    \label{fig:avg_session_duration}
\includegraphics[width=0.8\linewidth]{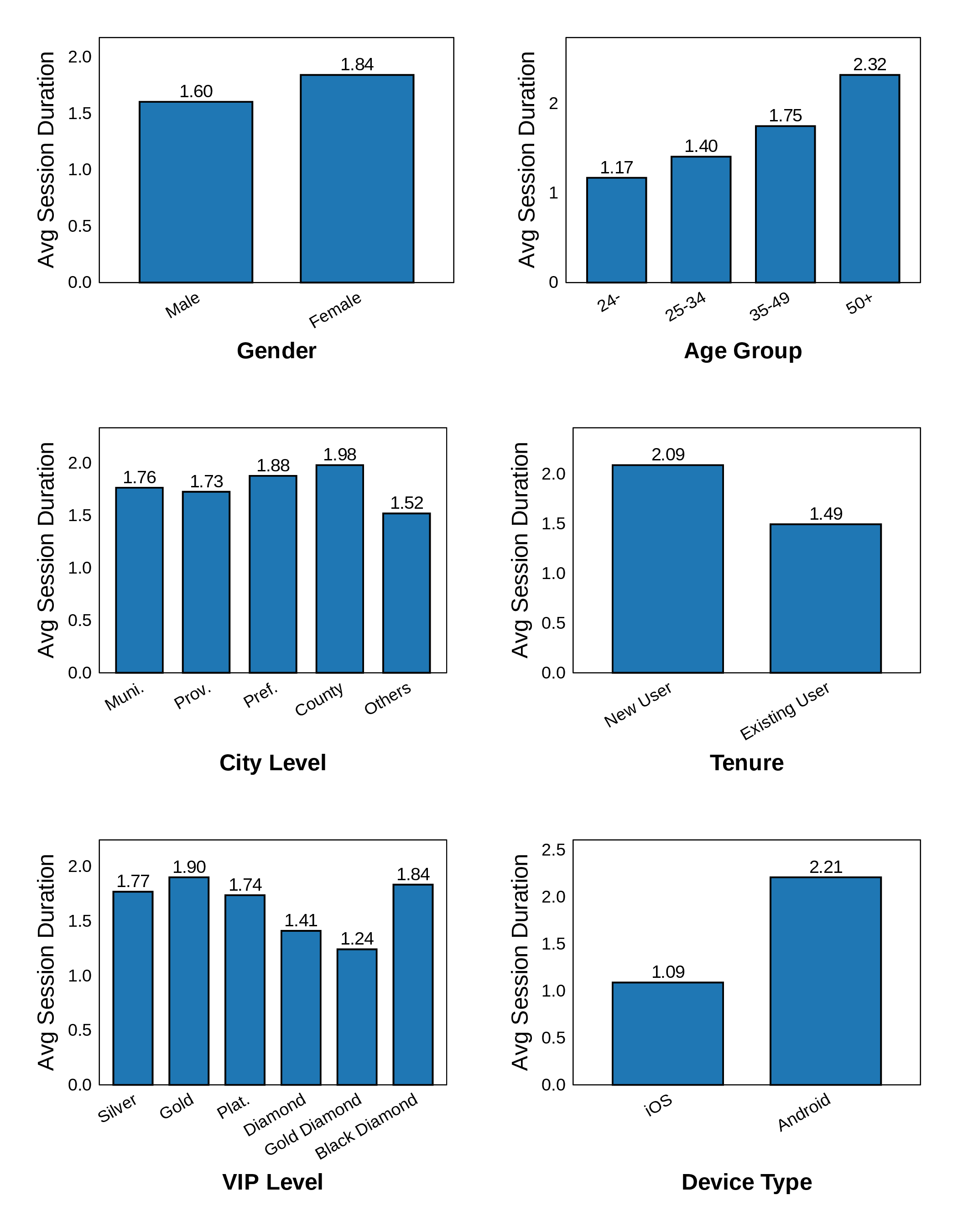}
    \begin{minipage}{0.8\linewidth}
    \setstretch{1}
    {\footnotesize \textit{Notes}: This figure shows the average AI chat session duration, measured in minutes, across six user characteristics: gender, age group, city level, tenure (new versus existing users), VIP level (Silver, Gold, Platinum, Diamond, Gold Diamond, and Black Diamond), and device type (Android versus iOS). Each panel reports the mean session duration for the categories within the corresponding characteristic. The averages are computed among users with at least one AI chat session during the sample period.}
    \end{minipage}
\end{figure}

\newpage

\begin{figure}[H]
\centering
    \caption{Average number of requests per AI chat session by demographic and account characteristics}
    \label{fig:avg_requests_per_session}
\includegraphics[width=0.8\linewidth]{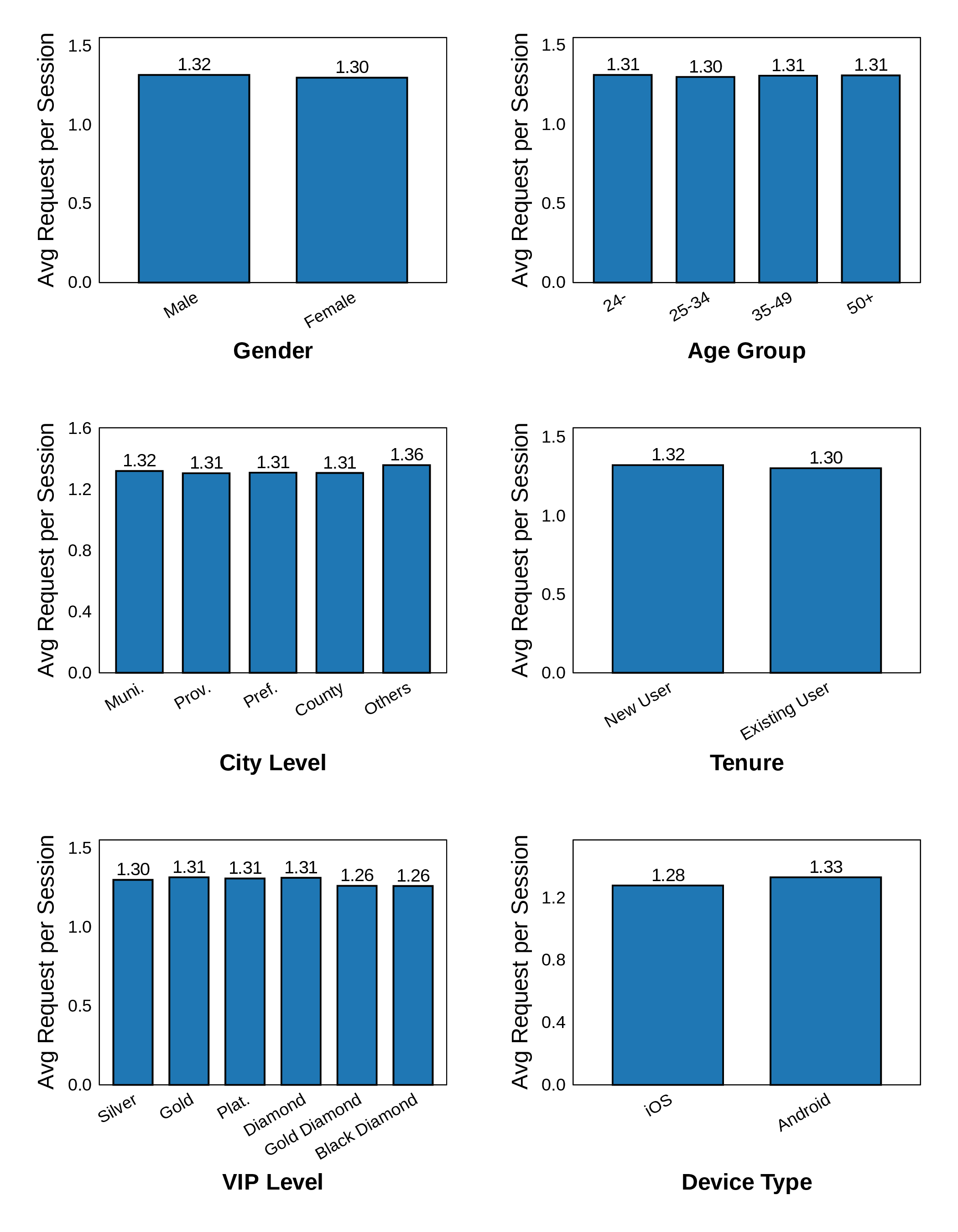}
    \begin{minipage}{0.8\linewidth}
    \setstretch{1}
    {\footnotesize \textit{Notes}: This figure shows the average number of user requests per AI chat session across six user characteristics: gender, age group, city level, tenure (new versus existing users), VIP level (Silver, Gold, Platinum, Diamond, Gold Diamond, and Black Diamond), and device type (Android versus iOS). Each panel reports the mean number of user requests per session for the categories within the corresponding characteristic. The averages are computed among users with at least one AI chat session during the sample period.}
    \end{minipage}
\end{figure}

\newpage

\begin{figure}[htbp!]
    \caption{Distribution of Chat Requests by Journey Progress Quintiles and Chat Intent}
\centering\includegraphics[width=0.9\linewidth]{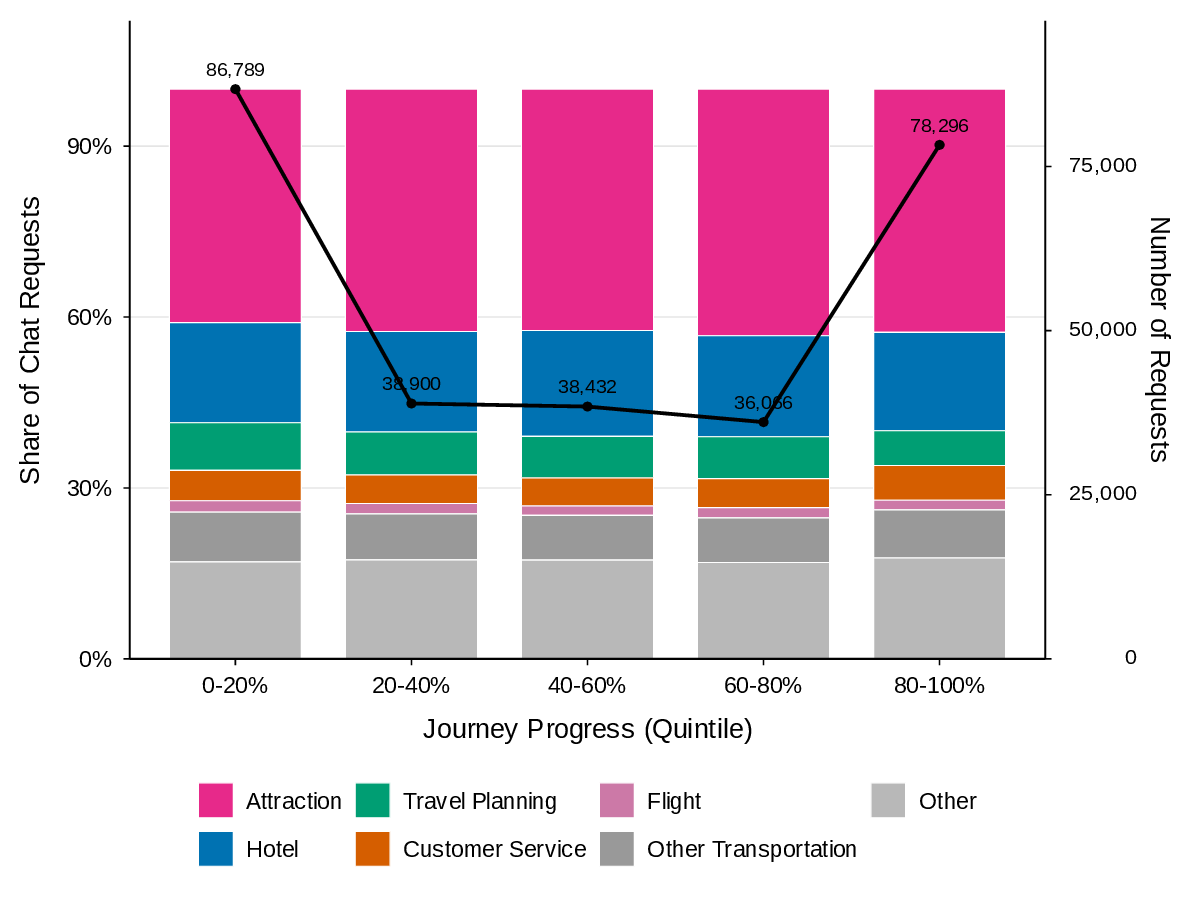}
    \label{fig:intent_by_progress}
    \begin{minipage}{0.9\linewidth}\setstretch{1}
    {\footnotesize \textit{Notes}: This figure shows the volume and intent composition of AI chat requests across quintiles of journey progress. The left axis and stacked bars report the within-quintile share of chat requests by intent category: Attraction, Hotel, Travel Planning, Consumer Support, Flight, Other Transportation, and Other. The right axis and overlaid line report the total number of chat requests in each quintile.}
    \end{minipage}
\end{figure}

\end{document}